\documentclass[runningheads]{llncs}

\usepackage{eccv}

\newcommand*{\img}[1]{%
    \hspace{-2\baselineskip} 
    \raisebox{-.7\baselineskip}{
        \includegraphics[
        height=1.8\baselineskip,
        width=1.8\baselineskip,
        keepaspectratio,
        ]{#1}%
    }%
}

\usepackage{eccvabbrv}

\usepackage{graphicx}
\usepackage{booktabs}

\usepackage[accsupp]{axessibility}  

\usepackage{hyperref}

\usepackage{orcidlink}

\begin{document}

\title{\img{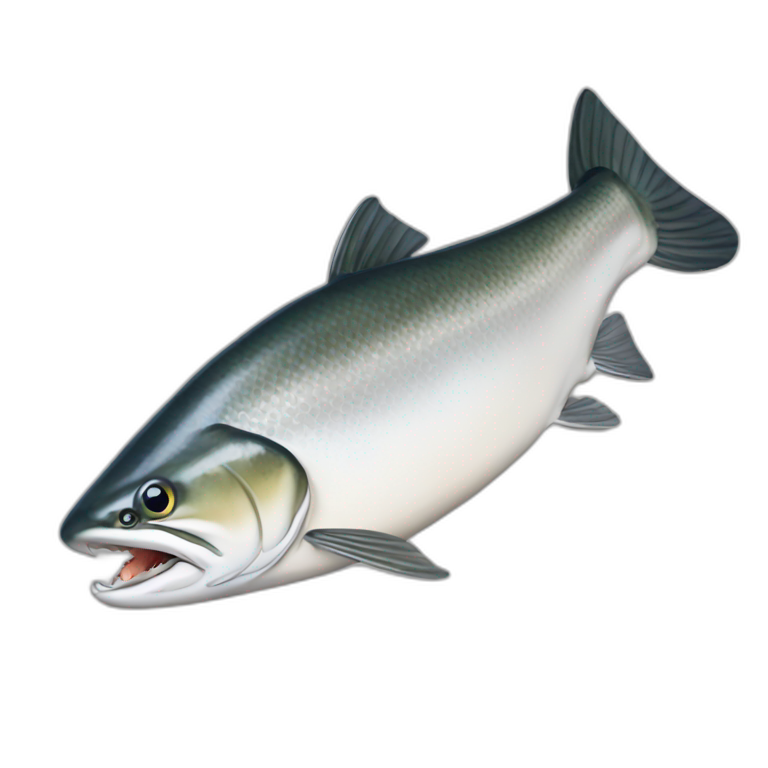}COHO: Context-Sensitive City-Scale Hierarchical Urban Layout Generation} 

\titlerunning{COHO: Context-Sensitive City-Scale Hierarchical Urban Layout Generation}

\author{Liu He\orcidlink{0000-0001-9715-2606},
Daniel Aliaga\orcidlink{0000-0001-9794-462X}}

\authorrunning{He, L. \& Aliaga, D.}

\institute{Purdue University, USA\\
\email{\{he425,aliaga\}@purdue.edu}
}

\maketitle

\begin{figure}[ht]
  \centering
  \includegraphics[width=0.85\linewidth]{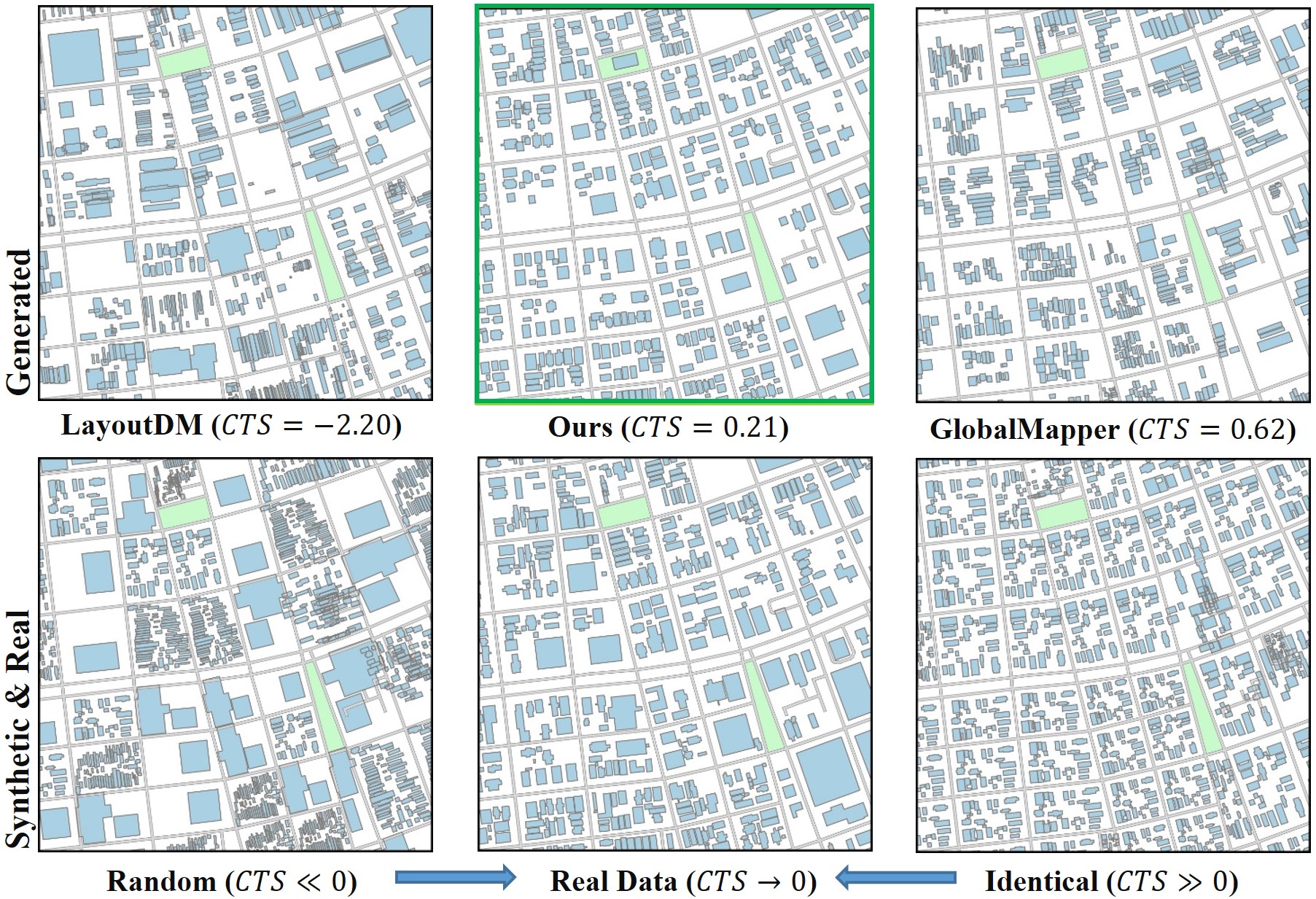}
  \caption{\textbf{Context-Sensitive Generation.} Our method (Green) pursues realistic context harmonization among neighboring city blocks as real data (bottom-middle). Other methods (e.g. LayoutDM~\cite{inoue2023layoutdm}, GlobalMapper~\cite{he2023globalmapper}) show over-diversity/-similarity in city-scale layout generation (evaluated by Context Score $CTS$ as Eq.~\ref{eqn:CTS2}.). Fully random and identical layouts are synthetically generated to illustrate extreme cases.}
  \label{fig:teaser}
\end{figure}

\begin{abstract}
The generation of large-scale urban layouts has garnered substantial interest across various disciplines. Prior methods have utilized procedural generation requiring manual rule coding or deep learning needing abundant data. However, prior approaches have not considered the context-sensitive nature of urban layout generation. Our approach addresses this gap by leveraging a canonical graph representation for the entire city, which facilitates scalability and captures the multi-layer semantics inherent in urban layouts. We introduce a novel graph-based masked autoencoder (GMAE) for city-scale urban layout generation. The method encodes attributed buildings, city blocks, communities and cities into a unified graph structure, enabling self-supervised masked training for graph autoencoder. Additionally, we employ scheduled iterative sampling for 2.5D layout generation, prioritizing the generation of important city blocks and buildings. Our approach achieves good realism, semantic consistency, and correctness across the heterogeneous urban styles in 330 US cities. 
Codes and datasets are released at~\url{https://github.com/Arking1995/COHO}.


\end{abstract}
\section{Introduction}
\label{sec:intro}
Large-scale urban layout generation has received significant interest by computer vision, urban planning and related disciplines. In particular, cities easily have 1,000 to 50,000 city blocks spanning a mostly horizontal terrain. The configuration of each city block varies significantly across the city, yet there is some configuration similarity amongst subsets within and among cities. Reconstructing existing layouts and generating future planned layouts are crucial tasks for urban simulation, digital twins, and game/content design. 

Prior works have generated urban layouts from a variety of sources. Procedural layout generation~\cite{lipp2011interactive, vanegas2012procedural} used hand-coded rules to generate cities, and inverse procedural modeling~\cite{bokeloh2010connection} extracted the rules from data and then enabled generation. More recently, deep learning has promoted data-driven approaches that produce either pixel-based layout generation (e.g., InfiniCity~\cite{lin2023infinicity}, CityDreamer~\cite{xie2023citydreamer}, CityGEN~\cite{deng2023citygen}) or graph-based approaches for general polygonal shape layouts (e.g., LayoutTransformer~\cite{gupta2021layouttransformer}, VTN~\cite{arroyo2021variational}) and for urban scenarios (e.g., BlockPlanner~\cite{xu2021}, GlobalMapper~\cite{he2023globalmapper}, BuildingGAN~\cite{chang2021building}). However, none of these techniques explicitly recover and consider the context-sensitive nature of urban layout generation (Fig.~\ref{fig:teaser}). In other words, the layout of buildings, blocks, and communities cannot be treated in isolation. The multi-layer semantics are critical to generating realistic and plausible urban layouts. While such multi-level structure (e.g., from single building to a community or subset of a city) has been specified via hand-coded rules, it has not been discovered, organized and used for generation in data-driven deep learning approaches.

Our approach builds on three key observations:
\begin{enumerate}
    \item \textbf{Representation:} Cities, communities, blocks, buildings, and roads are arbitrarily shaped and with complex topology. Hence, a graph-based representation can better capture the regular and non-regular configurations, as opposed to the limiting regular structure of an image. Further, graphs can be more compact which in turn supports better scalability. Altogether, having buildings or city blocks as atomic units, instead of pixels, will contribute to improving correctness, realism, and consistency with priors.
    
    \item \textbf{Context-Sensitivity:} Buildings, city blocks, and communities are built considering their neighboring structures and not in isolation. Hence, a generator should consider this multi-layer structure and inter-dependency. In order words, the style features at one level (e.g., size, shape, location) depend on the surrounding context. 
    
    \item \textbf{Prioritization:} The stylistic and semantic importance, or priority, of city blocks and buildings varies. This implies that a generation scheme should generate the most important city blocks and buildings first and then fill-in the rest, as opposed to a region growing approach for example. Overall, prioritization will improve correctness and realism as well as enable large-scale urban layout generation. 
    
\end{enumerate}

We propose a novel Graph-based Masked AutoEncoder (GMAE) for large-scale multi-layer 2.5D urban layout generation. \textit{First}, we define a canonical graph-based representation capturing the multi-layer context sensitivity of any city's urban layouts (Fig.~\ref{fig:graph_repr}). The entire graph $G$ corresponds to a city. Subsets of the graph map to communities and each node $b$ amounts to a city block. Further, each node encodes its building layouts, shapes, and heights as a quantized feature by well-trained quantizor. \textit{Second}, we utilize our GMAE to learn the multi-layer stylistic behavior of many urban layouts (i.e., 330 cities). 

Using subsets of the graph (or communities), node features are randomly masked and self-supervised GMAE training is performed to learn node feature reconstruction (Fig.~\ref{fig:model}). \textit{Third}, utilizing the well-trained GMAE as a generator, we perform an iterative priority-based scheduling to generate a large layout or to complete a city fragment; i.e., each iteration predicts node features for the entire graph but only the most confident ones are preserved for the next step (Fig.~\ref{fig:inference}). This methodology is able to leverage any percentage of given prior (e.g., $[0, 100\%]$) for urban layout completion and generation.

To the best of our knowledge, our approach is the first data-driven deep-learning method to enable large-scale 2.5D layout generation with good realism, semantic consistency, and correctness. We show the capability to generalize to the heterogeneous urban layout styles of 330 cities in the US (e.g., all cities with a population $>$100K, totaling 833,473 city blocks and 17,663,607 buildings). In addition, we provide comparisons to several prior works including SDXL~\cite{podell2023sdxl}), VTN~\cite{arroyo2021variational}, LayoutDM~\cite{inoue2023layoutdm}, and GlobalMapper~\cite{he2023globalmapper} in Sec.~\ref{sec:experiments} and exceed their performance in multiple well-known geometric and perceptual metrics.

Our contributions include the following:
\begin{itemize}
    \item[$\bullet$] A canonical graph representation for large-scale 2.5D urban layouts. It encodes global and local node features and neighboring relations for buildings, city blocks, and communities within a city.
    \item[$\bullet$] A self-supervised Graph-based Masked AutoEncoder (GMAE) enabling arbitrarily shaped city layout generation with consistency, correctness, and realism.
    \item[$\bullet$] Priority-based scheduled iterative sampling for controllable urban layout synthesis including completion, refinement, and generation with any percentage of priors ($[0, 100\%]$).
    \item[$\bullet$] A open dataset of road network and urban layouts of 330 US cities for benchmarking current and future urban layout generation methods (\url{https://huggingface.co/datasets/Arking95/COHO}).
\end{itemize}

\section{Related Work}
\label{sec: related works}

\subsubsection{Layout Generation.}
Numerous data-driven deep learning methods have been proposed for layout generation; for example, generalized layout synthesis~\cite{jyothi2019layoutvae, li2020layoutgan, gupta2021layouttransformer, arroyo2021variational, yang2021layouttransformer}, document layouts~\cite{patil2020read, hedocument2023, tabata2019automatic, jiang2023layoutformer++, zheng2019content}, urban land lots~\cite{xu2021,he2023generative,zhang2022guided}, indoor scenes~\cite{wu2019data, para2021generative, nauata2020house}, and 3D buildings~\cite{chang2021building, bhatt2020design,patel2023deep}. Recently, diffusion models have also been used for layout generation tasks (e.g., ~\cite{inoue2023layoutdm, hedocument2023, chai2023layoutdm, hui2023unifying}). However, most of these works assume 2D layouts generated on a rectangular empty canvas. The position and size of each bounding box is often limited to a finite set of categories and to rectilinear alignments. This results in poor diversity and low realism for urban layout generation tasks. BlockPlanner~\cite{xu2021} and GlobalMapper~\cite{he2023globalmapper} adapt layout generation to either rectilinear or arbitrarily-shaped city block generation. However, all previous works assume each layout, or city block, is independent. This is not the case in urban environments where adjacent and nearby city blocks may share a common style or be positioned in a purposeful multi-block arrangement. Without a contextual guidance, city block generation may result in either unwanted harmonization or unexpected diversity between neighboring blocks (see Fig.~\ref{fig:teaser}). Thus, we pursue city-scale urban layout generation with the help of context-aware neighboring information.
\smallskip
\subsubsection{Infinite Visual Synthesis.}
With bombing of visual deep learning works~\cite{song2023objectstitch,song2024imprint,sheng2022controllable,hua2024finematch, sheng2024dr,ma2022learning}, researchers have explored synthesizing large, potentially unbounded images. For example, omnidirectional image synthesis from multiple smaller field-of-view images~\cite{lin2021infinitygan, li2022infinitenature}. Other works focus on unbounded 3D scene generation~\cite{chai2023persistent, shen2022sgam, chen2023scenedreamer}. Particularly relevant are several recent infinite urban scene generators~\cite{lin2023infinicity, deng2023citygen, xie2023citydreamer} that directly adapt image synthesis methods (e.g. InfinityGAN~\cite{lin2021infinitygan}, MaskGIT~\cite{chang2022maskgit}). The synthesis performs an autoregressive outpainting process using layout image patches. However, patch-based training and inference doesn't learn global features nor their interdependencies. Rather, it learns a very local behavior. Prior pixel-based methods acknowledge limited realism and semantics ~\cite{lin2023infinicity, deng2023citygen, xie2023citydreamer}. Often, post-processing~\cite{deng2023citygen, xie2023citydreamer}, and human-in-the-loop editing~\cite{lin2023infinicity} is required to refine the generated images to vector-based urban layouts. Even the latest stable diffusion model SDXL~\cite{podell2023sdxl} seeks high-resolution image synthesis but may not produce realistic urban layouts (Fig.~\ref{fig:comparison}).

\smallskip
\subsubsection{Learning based on Quantized Representations.}
\label{sec:related_learning_quantized}
Vector quantization has been a prevailing stepping-stone technique for text (e.g., Word2Vec~\cite{mikolov2013efficient}), image (e.g., VQVAE/GAN~\cite{van2017neural, esser2021taming}, ~\cite{dosovitskiy2020image}), and video synthesis (e.g., SORA~\cite{videoworldsimulators2024, peebles2023scalable}, ~\cite{yan2021videogpt}). Pretrained quantizers provide a scalable token representation benefiting high-level and large-scale synthesis. A well-trained quantizer combined with self-supervised masked training often performs well in reconstructing the original signal during generative tasks. BERT~\cite{devlin2018bert} provides plausible text-based representation learning. MAE~\cite{he2022masked} defines a masked autoencoder for image space which excels at multiple downstream tasks. MAE also provides representation learning for a broad range of node/graph feature synthesis~\cite{hou2022graphmae, hou2023graphmae2}. Specifically, MaskGIT~\cite{chang2022maskgit} and MAGE~\cite{li2023mage} utilize MAE frameworks for impressive image generation tasks. The idea of a generative masked autoencoder based on pretrained quantization was inspirational to our city-scale urban layout approach.

\section{Method}
\label{sec:method}
First we describe our canonical graph representation of 2.5D urban layouts (Fig.~\ref{fig:graph_repr}). Second, we explain the training process for our GMAE based on the graphs for many cities (Fig.~\ref{fig:model}). Third, we show our priority-based scheduling for iteratively generating city-scale urban layouts given any amount of prior (Fig.~\ref{fig:inference}).

\subsection{Canonical Graph Representation}
\label{sec:canonical_repr}

\begin{figure}[tb]
  \centering
  \includegraphics[width=0.9\linewidth]{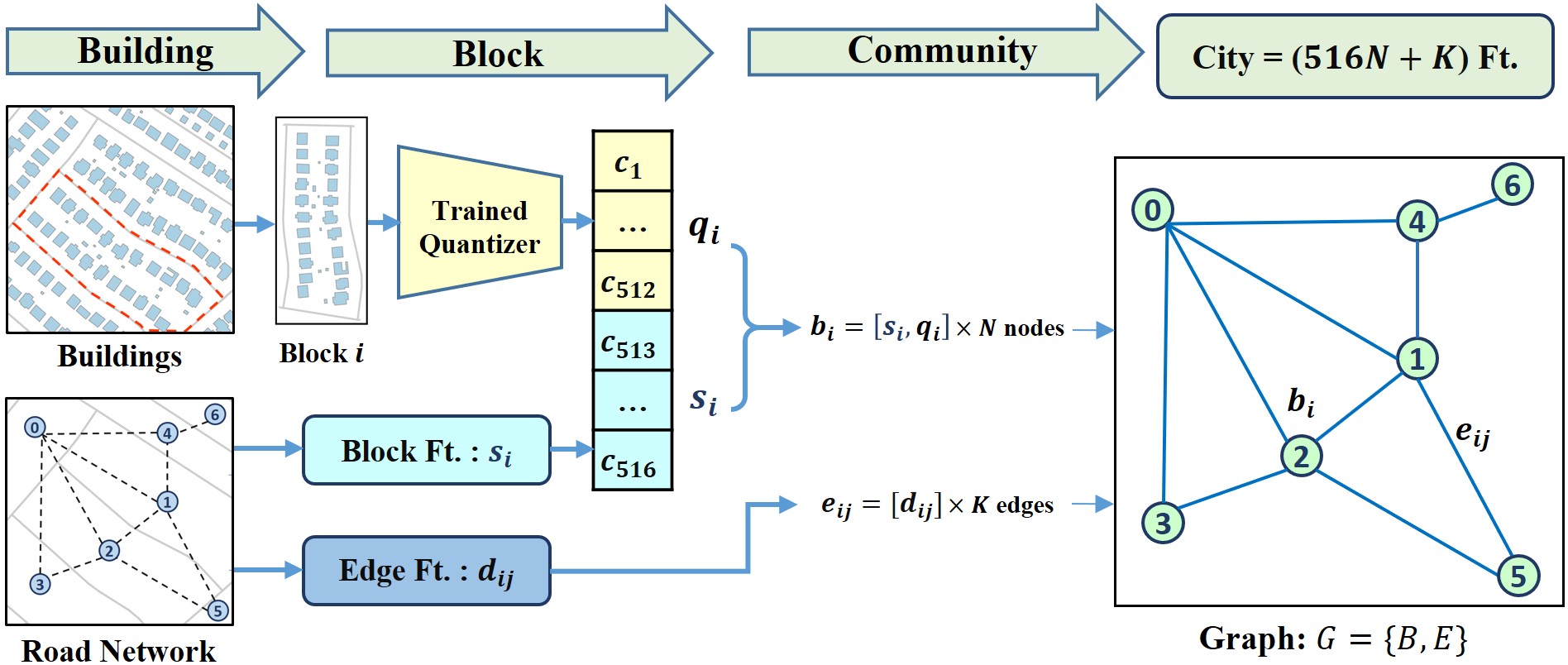}
  \caption{\textbf{Canonical Graph Representation.} Our method represents a city as a canonical graph $G$. Each node $b_{i}$ represent a single city block, and each edge $e_{ij}$ connects spatially adjacent blocks. Each block/node corresponds to a set of node features $s_{i}$ and a quantized vector $q_{i}$ hierarchically capturing enclosed building layouts. Edge feature $d_{ij}$ encodes distances between block centroids. Graph $G$ is used for GMAE training.}
  \label{fig:graph_repr}
\end{figure}

A city is represented by a graph $G = \{B, E\}$, where $B$ is a set of nodes $B = \{b_{i}| i\in [1, N]\}$ such that $b_{i}$ represents each city block (Fig.~\ref{fig:graph_repr}), and any connected subgraph of $G$ corresponds to a community. $N=|B|$ is the total number of blocks for a given city. $E$ is a set of edges $E = \{e_{ij} | i, j \in [1, N]\}$ representing mutually adjacent nodes (or city blocks). $K=|E|$ is the total number of edges.

\subsubsection{Node (or City Block).} Each node, or city block, $b_{i}$ is a region defined by an enclosing road network and contains a set of block-level features and building-level features. The per-node features include:

\begin{itemize}
    \item[$\bullet$] \textbf{$s_{i}$} is the shape and location feature vector of a city block contour. It includes four features: (1) aspect ratio, (2) total block area, (3) ratio of total block area over the block's convex-hull area, and (4) relative distance from current block to the centroid of the city.
    \item[$\bullet$] \textbf{$q_{i}$} is a 512-dimensional vector from a codebook $C$. It hierarchically captures the buildings layouts and their configuration within $b_{i}$.
\end{itemize}

We concatenate these node features to form a 516-dimensional vector ($[s_{i}, q_{i}]$), which is able to represent our large and heterogeneous set of blocks and buildings.

\subsubsection{Building Layout Quantization.}
The 512-dimensional vector $q_{i}$ describing the building layouts within a city block is obtained from a quantized codebook $C=\{1, 2, ..., L\}$ with quantization level $L=|C|$. To create the codebook, we train a block-level variational autoencoder (BVAE) for building layout within a city block, using a canonical graph structure for representing the multiple building shapes, positions, and heights inside a single city block. Using this pretrained encoder, the latent vector of all possible building layouts will form a numerical distribution for each latent dimension. We further quantize each of these distributions into $L$ bins with equal percentiles. Then, each latent value is replaced by the rank of the bin containing its value. Therefore, any building layout within a city block is quantized into a 512-dimensional categorical index vector $q_{i}=[c_{1}, ..., c_{512}]$, where each $c\in C$. Note that our codebook is defined after BVAE training and not dynamically trainable as in VQGAN/VQVAE~\cite{van2017neural,esser2021taming}. After performing a set of detailed evaluations using different training schemes, encoding models, and quantization levels $L$ (see Sec.~\ref{sec:ablation}), we selected the Graph Attention Network (GAT)~\cite{velickovic2017graph} as the backbone for the BVAE. Further, we select $L=20$ to balance compactness and representation ability after quantization. BVAE and quantization details are found in Supplementary Materials.

\subsubsection{Edge.}
Each edge $e_{ij}$ contains a feature $d_{ij}$ storing the relative distance between the centroids of the adjacent city blocks $b_{i}$ and $b_{j}$. This distance-weighted adjacency helps encode the relationship between neighboring blocks.

Altogether, the shape, style, and multi-layer urban layout of a city with $N$ nodes and $K$ edges is represented by a total of $(516N + K)$ variables. This canonical graph is used for the self-supervised GMAE training described below.

\subsection{Graph-based Masked AutoEncoder}

\begin{figure}[tb]
  \centering
  \includegraphics[width=\linewidth]{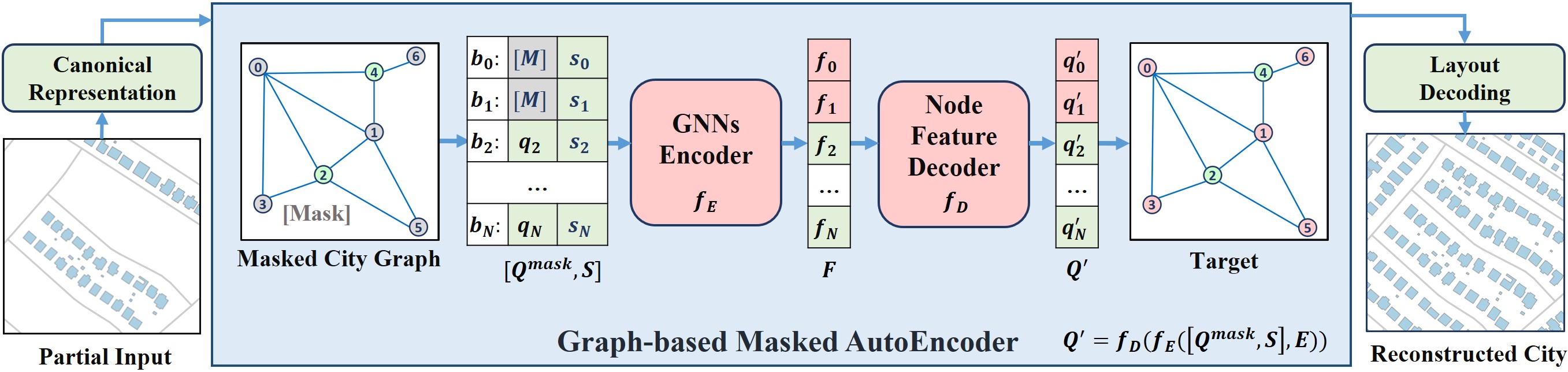}
  \caption{\textbf{Graph-based Masked Autoencoder.} Given a canonical city graph, quantized building layout features $Q$ are masked with dynamic masking ratios $m\in[0.5, 1.0]$, while block shape and location features $S$ are kept. The GNN encoder uses message passing between neighboring nodes to obtain the context-aware node features $F$. The decoder uses $F$ to reconstruct $Q'$. The predicted $Q'$ are decoded to 2.5D urban layouts.
  }
  \label{fig:model}
\end{figure}

We train our GMAE using a self-supervised process. A road-only city (or community) is a graph (or subgraph) $G$ where building layout features $q_{i}$ of every node (block) are missing, but $s_{i}$ and $e_{ij}$ are preserved. A context-sensitive node feature generation process corresponds to recreating the purposefully removed building layout features.

During training (Fig.~\ref{fig:model}), we use dynamic masking ratios of $q_{i}$ in order to learn how to regenerate the original features. In each training iteration, the mask ratio $m$ is sampled from a truncated Gaussian distribution as recommended by Li et al.~\cite{li2023mage}. The ratio is obtained from the range $m\in[0.5, 1.0]$ and is centered at $0.55$. We denote building layout features as $Q=\{q_{i}|i\in[1, N]\}$ and those after masking as $Q^{mask}$. Block features are denoted as $S = \{s_{i}|i\in[1, N]\}$. The graph neural network (GNN) encoder is denoted as $f_{E}$, and node feature decoder as $f_{D}$. Thus the process of our GMAE can be written as:

\begin{equation}
\label{eqn:objective}
F=f_{E}([Q^{mask}, S], E), \;\;\;  Q'=f_{D}(F)
\end{equation}

where $F=\{f_{i}| i\in[1, N]\}$ indicates the context-sensitive node features obtained by message passing between neighboring nodes by $f_{E}$. Extracted $F$ is processed by $f_{D}$ to reconstruct building layout features $Q'$. 

The backbones of $f_{E}$ and $f_{D}$ can be any GNN, including GAT~\cite{velivckovic2017graph}, GCN~\cite{kipf2016semi}, and GraphSAGE~\cite{hamilton2017inductive}. After experimentation, we select GAT as $f_{E}$ because of its best context-sensitive feature extraction ability, and we choose a straightforward MLP as $f_{D}$ for efficient processing. We observed that the re-masking trick and complex GNN decoder utilized by~\cite{hou2022graphmae, hou2023graphmae2} requires more memory and calculation but has no net benefit in our application. Since $Q$ contains quantized index vectors, we use cross entropy loss for self-supervised reconstruction training on masked nodes. Detailed ablations of masking strategy and model selections are provided in Sec.~\ref{sec:ablation}. 

Our objective function is as follows, where $L$ indicates the length of the quantized codebook $C$:

\begin{equation}
\label{eqn:rec}
L_{recon}= - {\sum_{i=1}^{L}} [{Q_{i}} {\log(Q'_{i})}]^{mask}
\end{equation}

To capture community behavior (and also reduce GPU memory usage), we randomly sample a batch of subgraphs from $G$ for each training iteration. The nodes in each subgraph are sampled using a chosen radius from a random center node. While community sizes vary significantly, we found that a practical compromise is to use 500 meters as the radius value. This radius ensures to cover a reasonable number of the city blocks with a typical census tract from governmental CJEST dataset~\cite{cjest}.

\begin{figure}[tb]
  \centering
  \includegraphics[width=\linewidth]{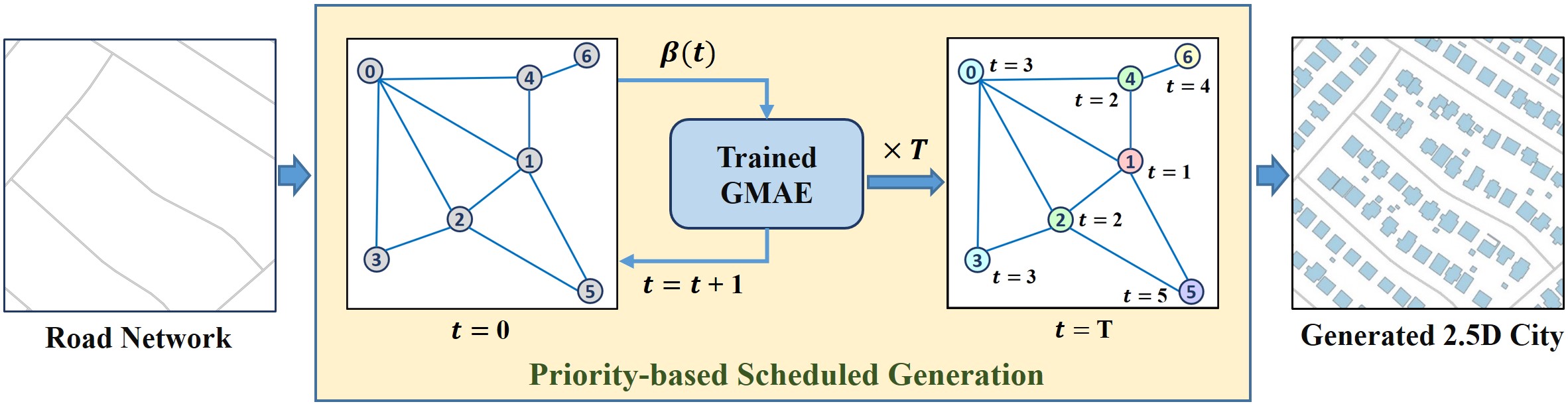}
  \caption{\textbf{Priority-based Scheduled Generation.} We iteratively utilize pretrained GMAE to reconstruct masked node features. In each iteration, we accept a certain ratio of predicted nodes decided by the scheduling function $\beta(t) = 1-cos(t/T)$. We obtain a full graph after $T$ iterations.
  }
  \label{fig:inference}
\end{figure}

\subsection{Priority-based Scheduled Generation}
\label{sec:iterative_sample}
While in theory the trained GMAE can predict all nodes in a single pass, the generation quality is much better when performed iteratively. Recent published papers (e.g., ~\cite{chang2022maskgit, li2023mage}) have found that gradually sampling with a reasonable number of iterations improves generation quality. In urban scenarios, small number of representative city blocks are important to indicate context behavior. We intend to generate those blocks at early iterations and then span neighboring blocks.

In Fig.~\ref{fig:inference}, we propose a $T$-iteration sampling strategy similar to MaskGIT~\cite{chang2022maskgit} and MAGE~\cite{li2023mage}. At each iteration, GMAE predicts per-node building layout features for all remaining nodes. Then we rank generated nodes by their prediction confidence. The generated nodes with the highest prediction confidence are accepted. The accepted nodes after the current iteration $t\in[1, T]$ are scheduled by a cosine function $\beta(t) = 1-cos(t/T)$. When $t$ is small, the ratio is near to 0 causing few nodes to be accepted in initial iterations. As iterations continue, $t$ gets bigger and hence the number of accepted nodes during each iteration will also increase. This behavior follows the prioritization observation in Sec.~\ref{sec:intro} which causes distinguished, or unique, city blocks to be generated first and essentially guide the layout style for nearby city blocks. Typically we set $T=12$ which results in dozens of nodes being generated as is the typical size of a community. Detailed ablations of alternative sampling schedules are provided in Sec.~\ref{sec:ablation}.

\section{Experiments}
\label{sec:experiments}

\subsection{Implementation Details}

\subsubsection{Open Dataset}
We have created a vector-based urban layout dataset of all US cities with a population of 100K or more. \textit{We believe this dataset will be of significant use to the community to further develop urban layout generation methods and thus we will release the dataset together with the paper}. For each city, we define a rectangular bounding box containing most of the metropolitan area. Then, we use this bounding box to extract data from OpenStreetMap (OSM)~\cite{OpenStreetMap}, Microsoft Building Footprints (MSF)~\cite{heris2020rasterized}, and Topologically Integrated Geographic Encoding and Referencing (TIGER) dataset~\cite{tiger}. The data we collected includes: (1) road network vectors and attributes, (2) building footprints and heights, and (3) block-level social economic metrics. Specifically, we collect road networks and block contours from TIGER. Then we look for building footprints and heights from OSM and MSF within each city block contour. We choose the one with more and accurate details (e.g., containing more buildings, having features that match those in TIGER). Over all 330 cities, this dataset has a total of 833,473 city blocks, and 17,663,607 buildings. We set training, validation, testing split as 70\%, 20\%, and 10\% respectively.

\subsubsection{Training Details}
The training has two stages. We first use all city blocks for self-supervised training of a graph-based Block-level Variational AutoEncoder (BVAE). Given this trained block-level encoder, we perform the latent quantization for all building layouts in each city block. Then our GMAE is trained based on canonical city graph representation. Training time for BVAE and GMAE typically takes 12 hours and 15 hour respectively by a single A5000 GPU. We will release our codes upon acceptance.

Inferencing using our graph-based approach is seemingly instantaneous for all the results shown in this paper. This is contrast to the VTN~\cite{arroyo2021variational} and SDXL~\cite{podell2023sdxl} approaches which take significantly more time to infer and to train.

\begin{figure}
\centering
\setlength{\tabcolsep}{1.5pt}
\begin{tabular}{ccccc}

    \includegraphics[width=0.19\textwidth]{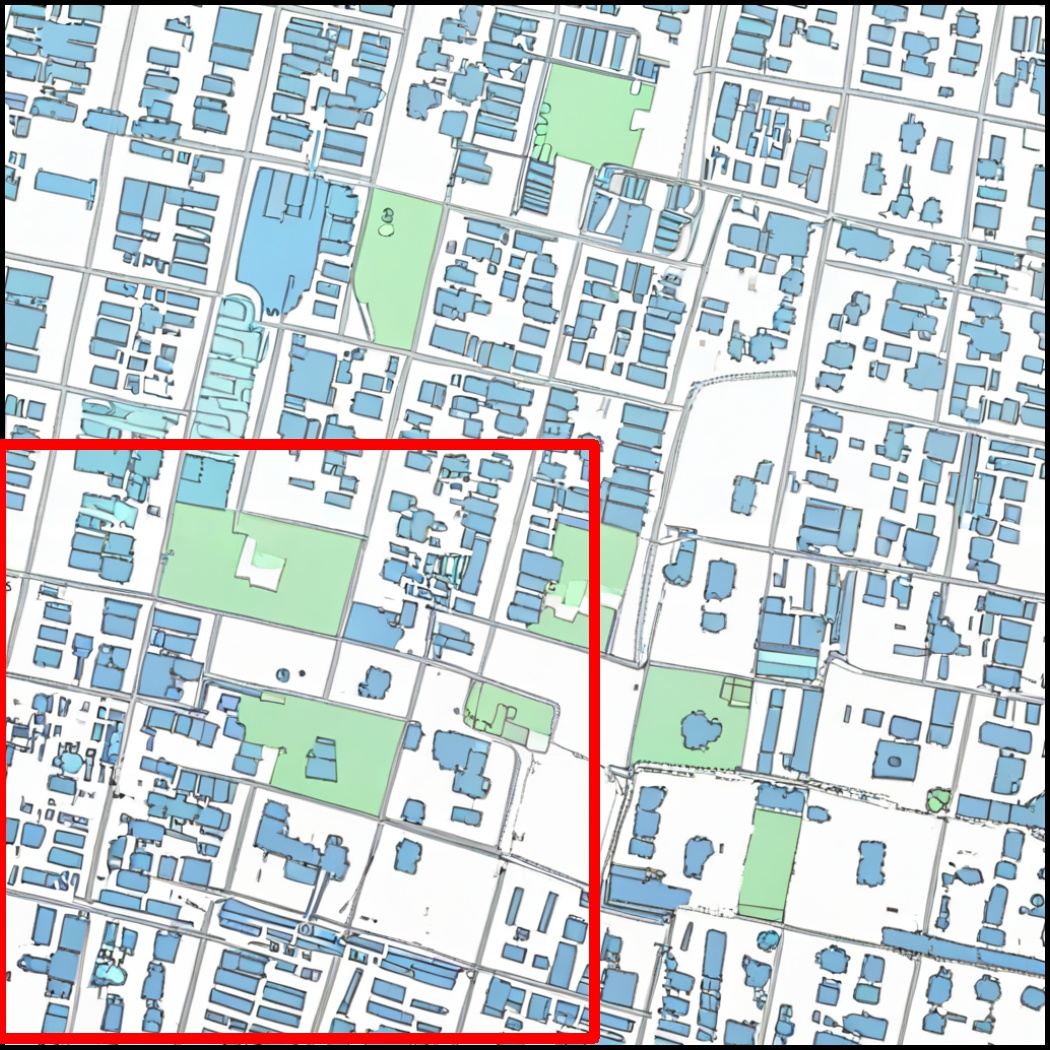} &
    \includegraphics[width=0.19\textwidth]{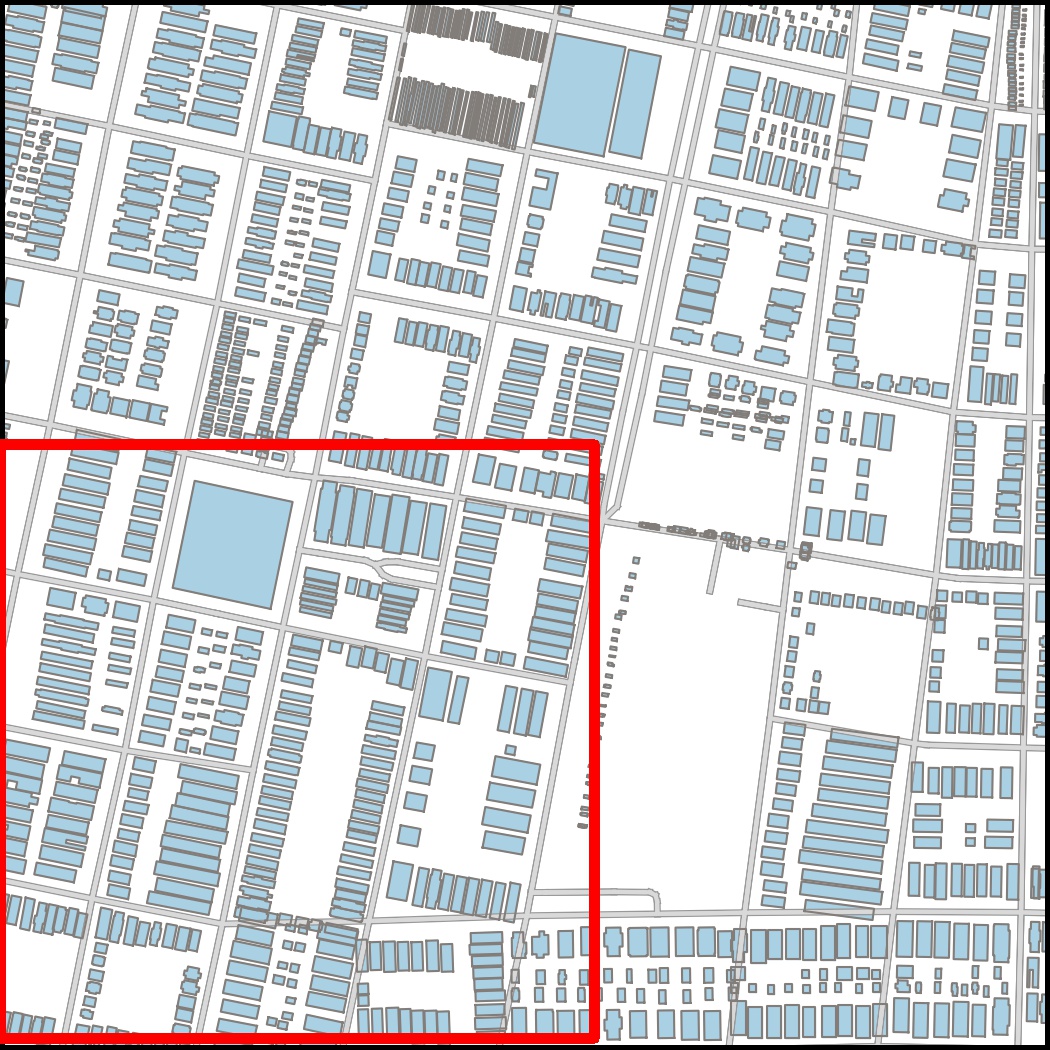} &
    \includegraphics[width=0.19\textwidth]{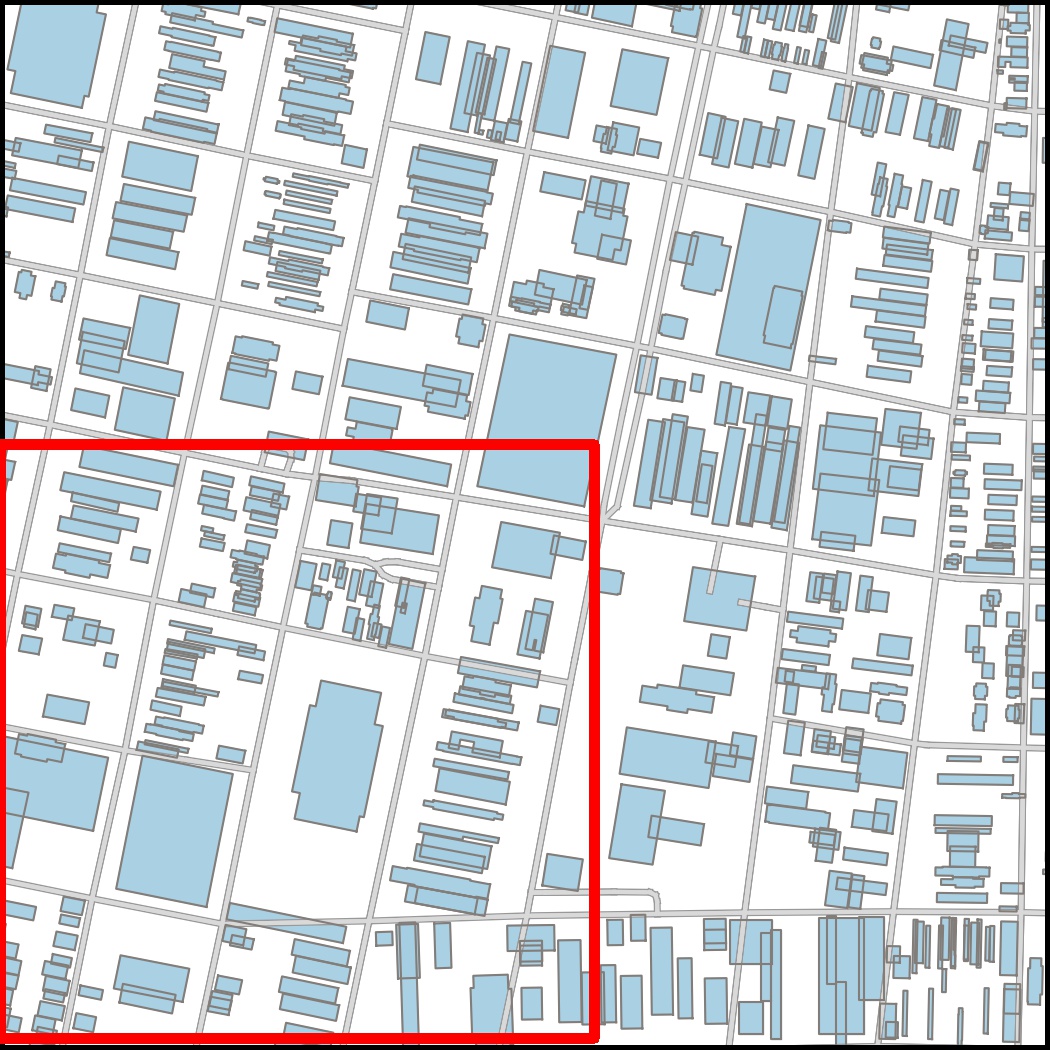} &
    \includegraphics[width=0.19\textwidth]{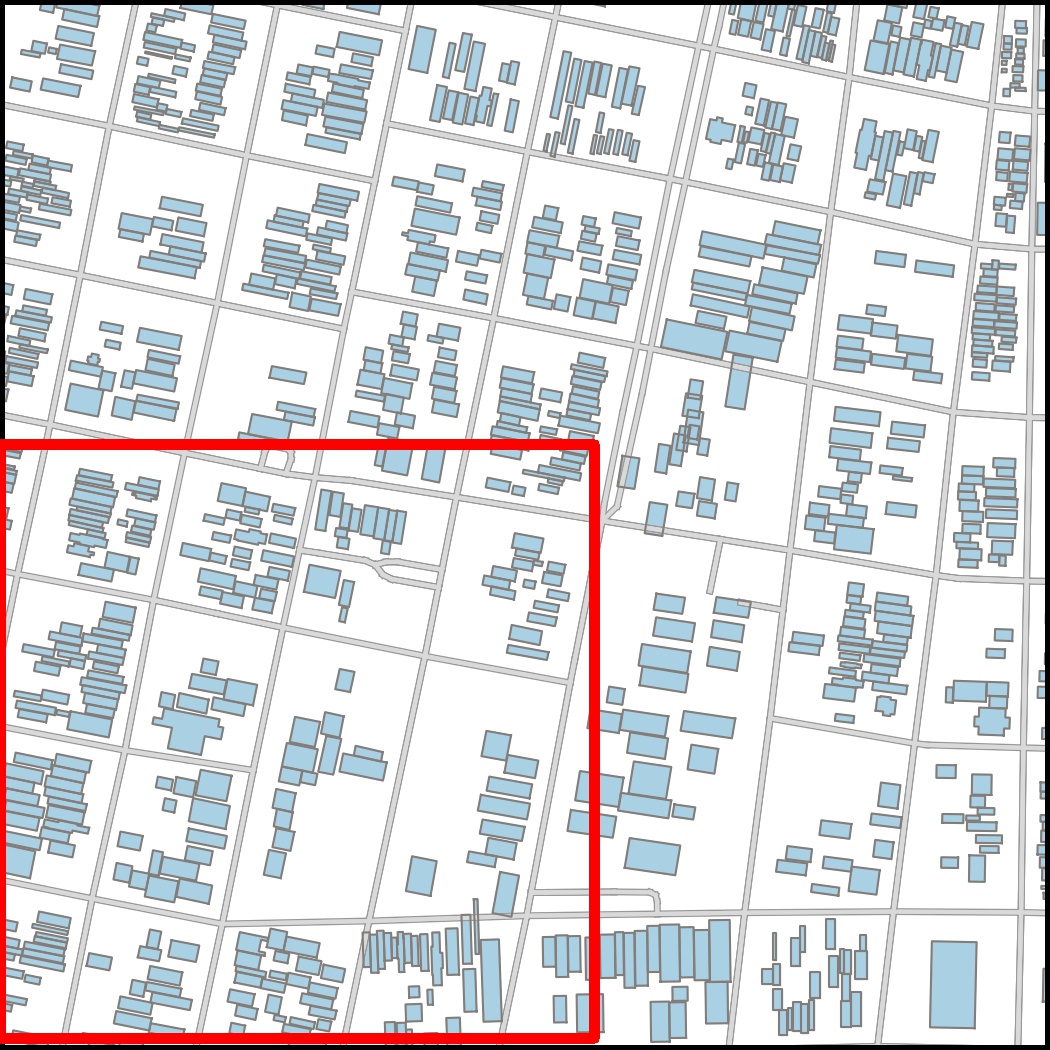} &
    \includegraphics[width=0.19\textwidth]{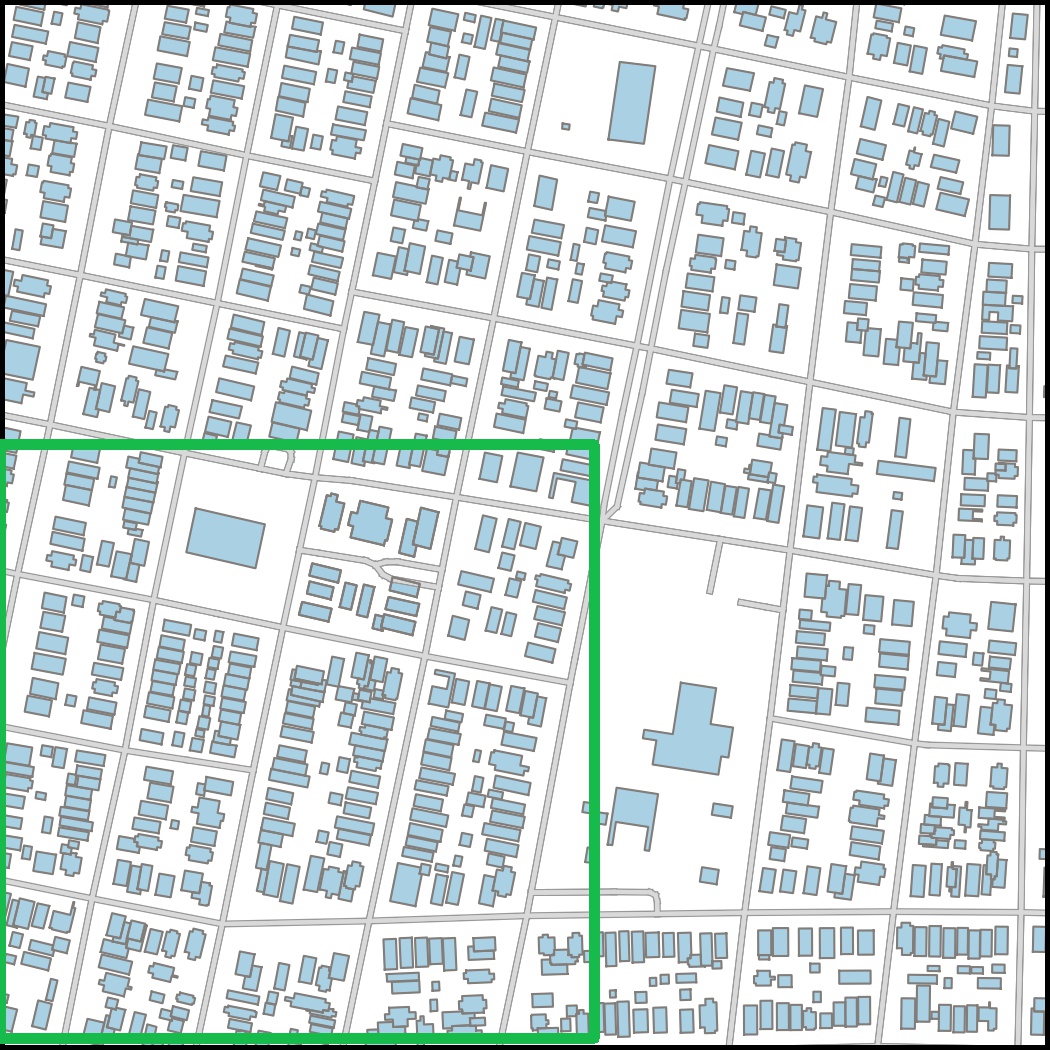} \\


   \includegraphics[width=0.19\textwidth]{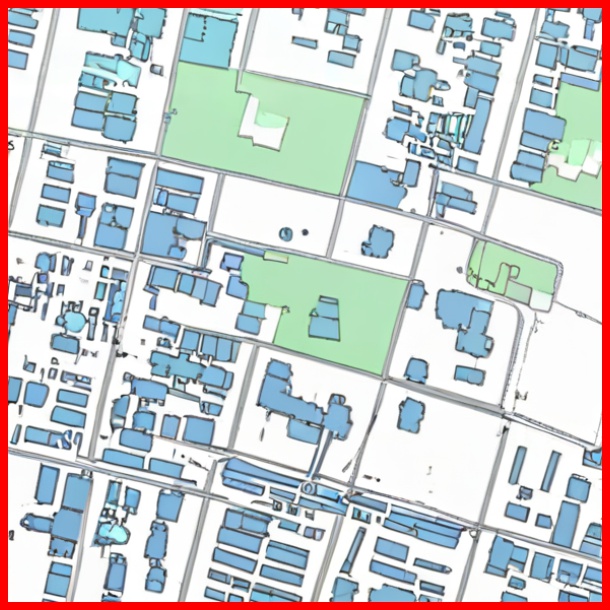} &
   \includegraphics[width=0.19\textwidth]{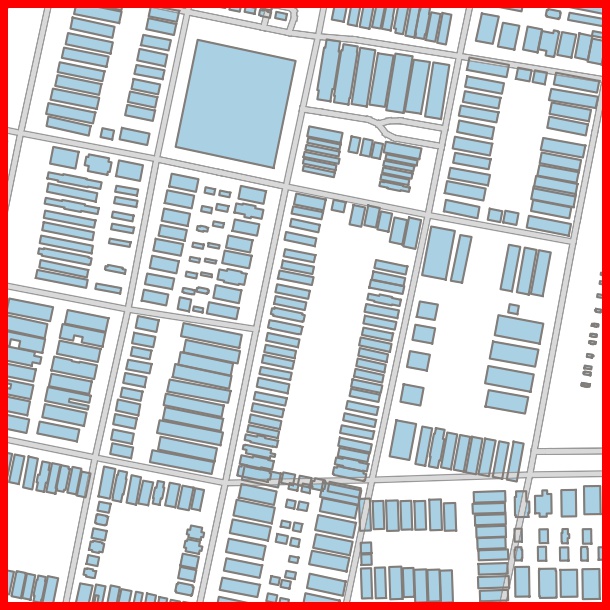} &
   \includegraphics[width=0.19\textwidth]{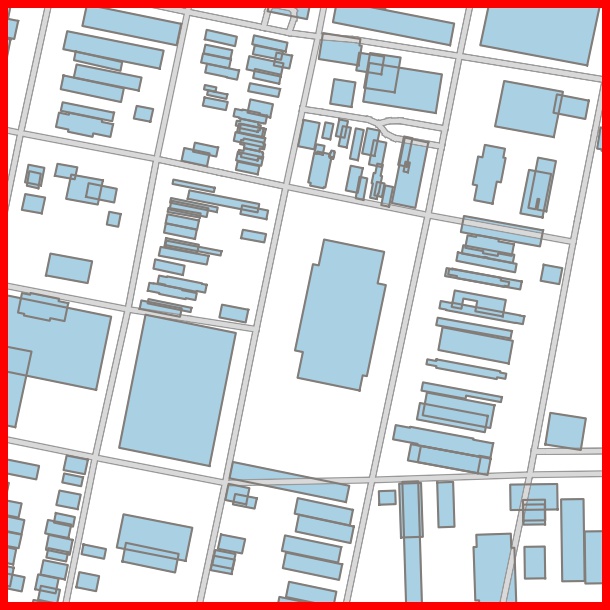} &
   \includegraphics[width=0.19\textwidth]{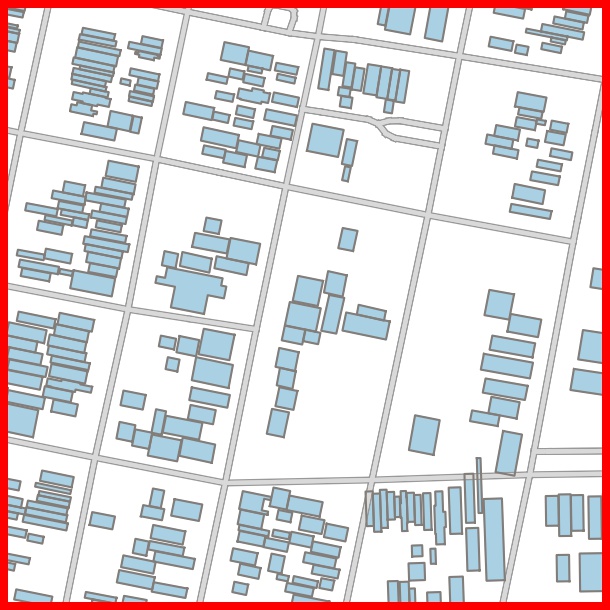} &
   \includegraphics[width=0.19\textwidth]{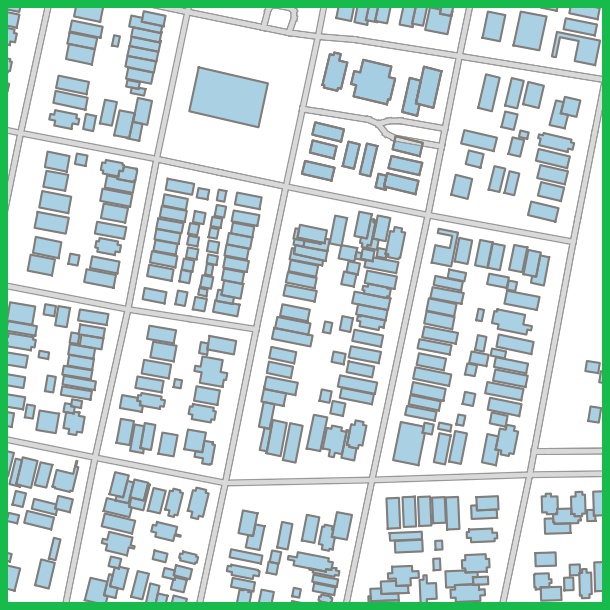} \\

    \includegraphics[width=0.19\textwidth]{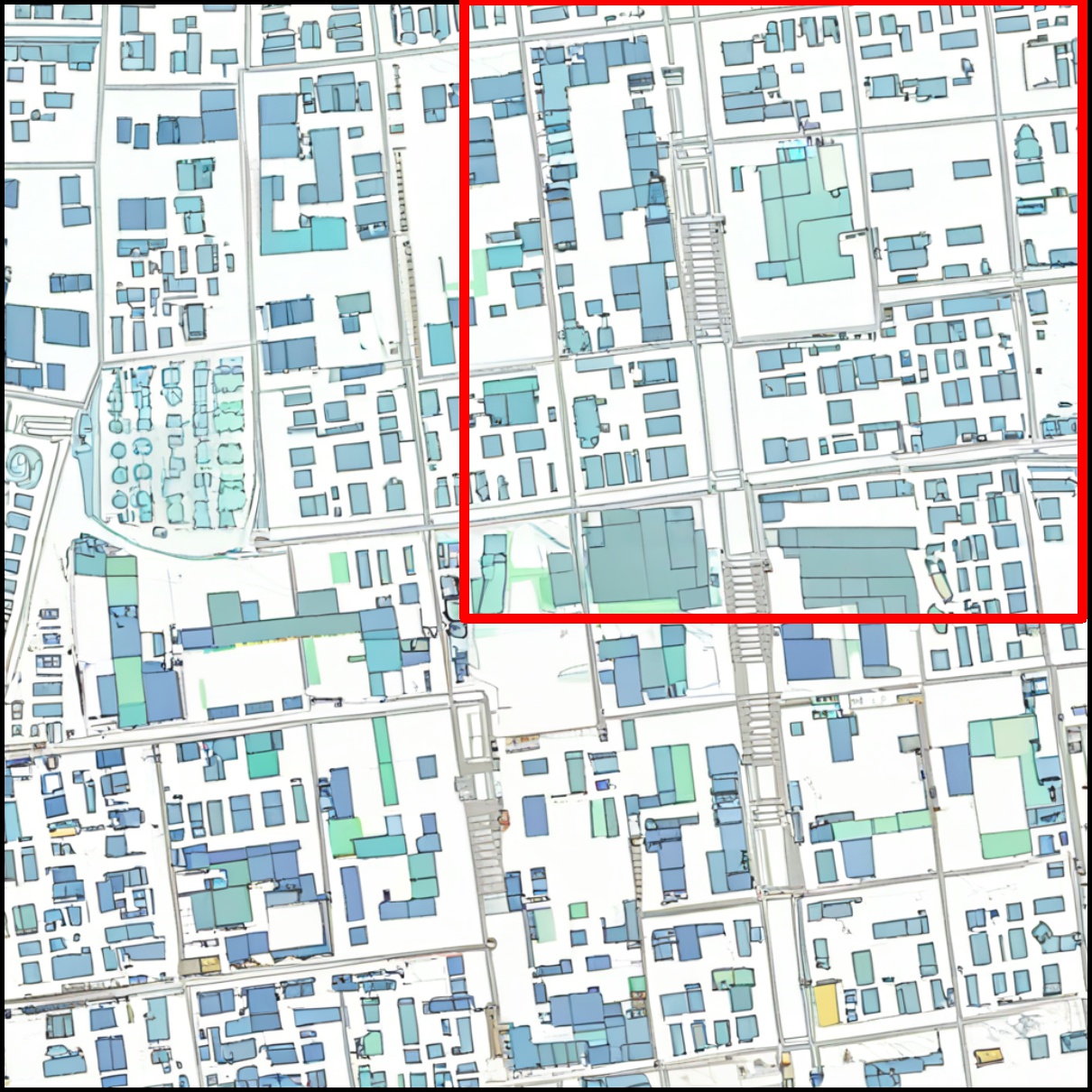} &
    \includegraphics[width=0.19\textwidth]{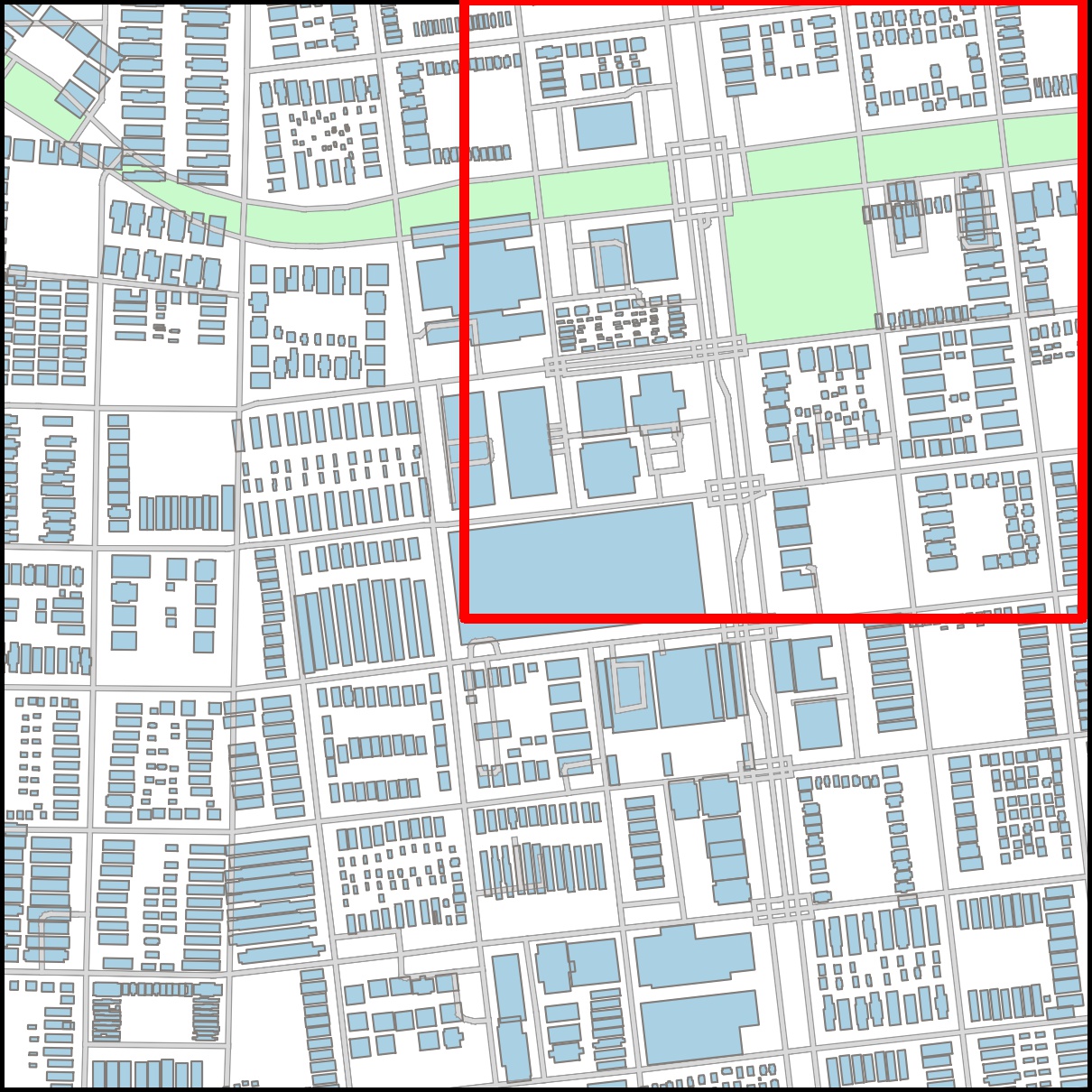} &
    \includegraphics[width=0.19\textwidth]{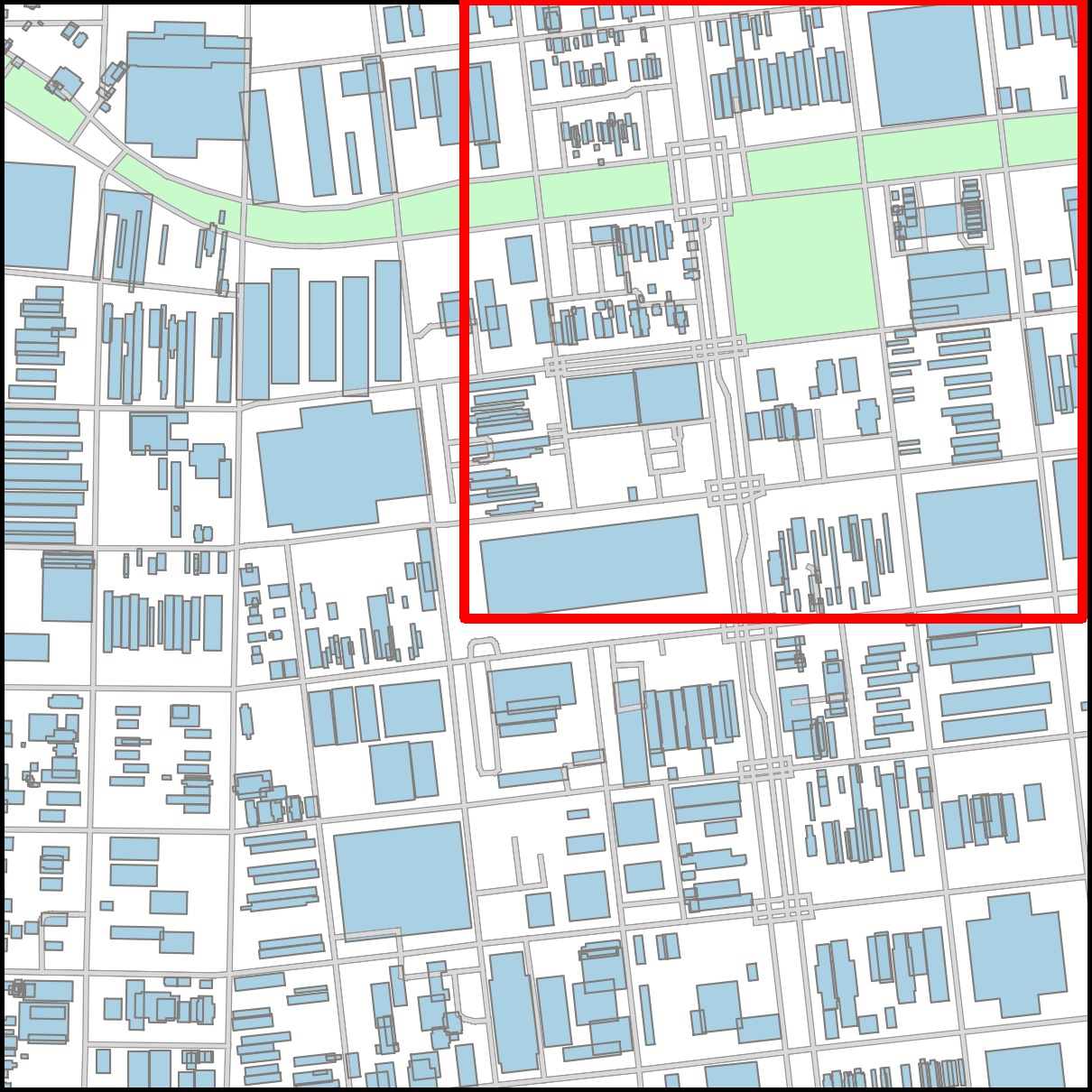} &
    \includegraphics[width=0.19\textwidth]{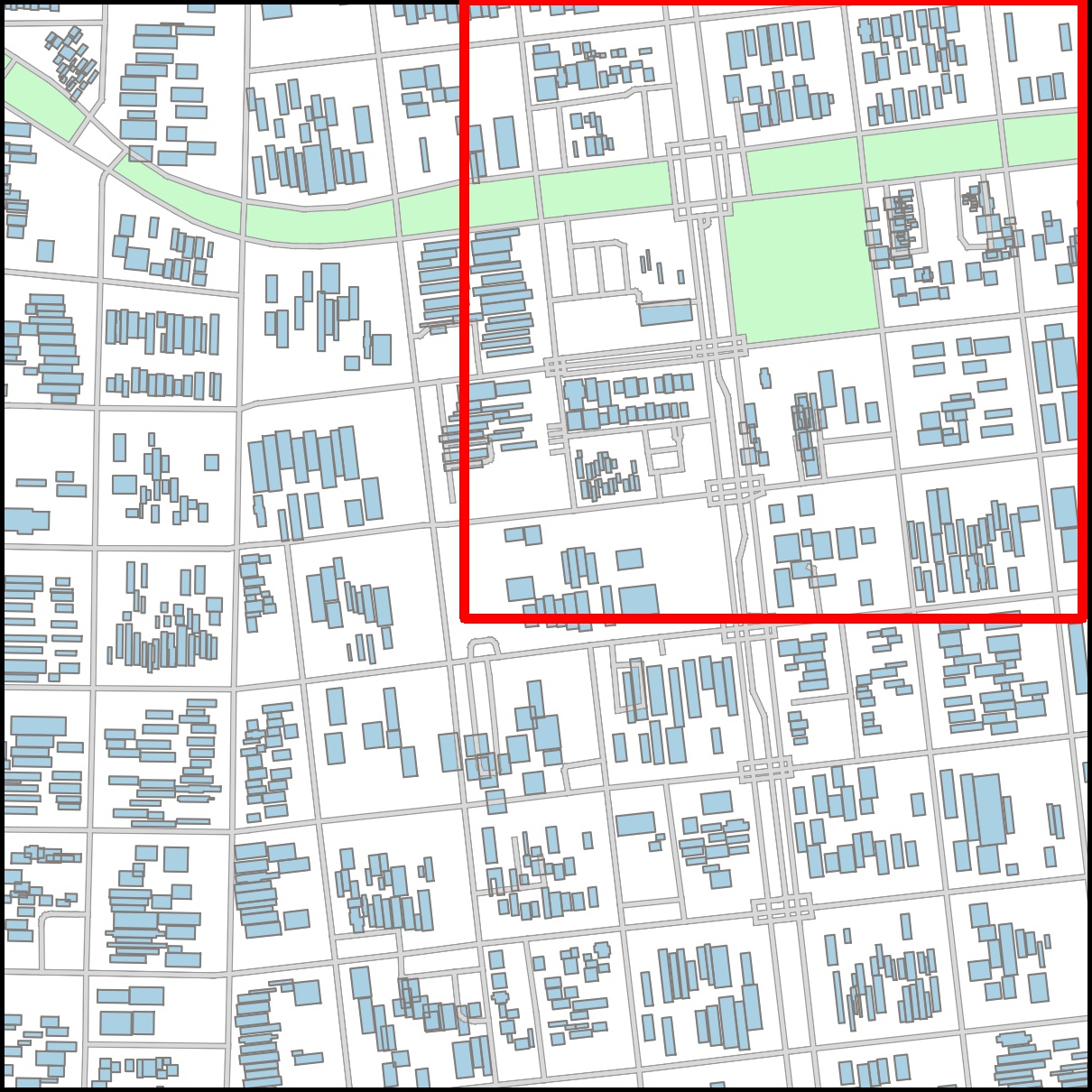} &
    \includegraphics[width=0.19\textwidth]{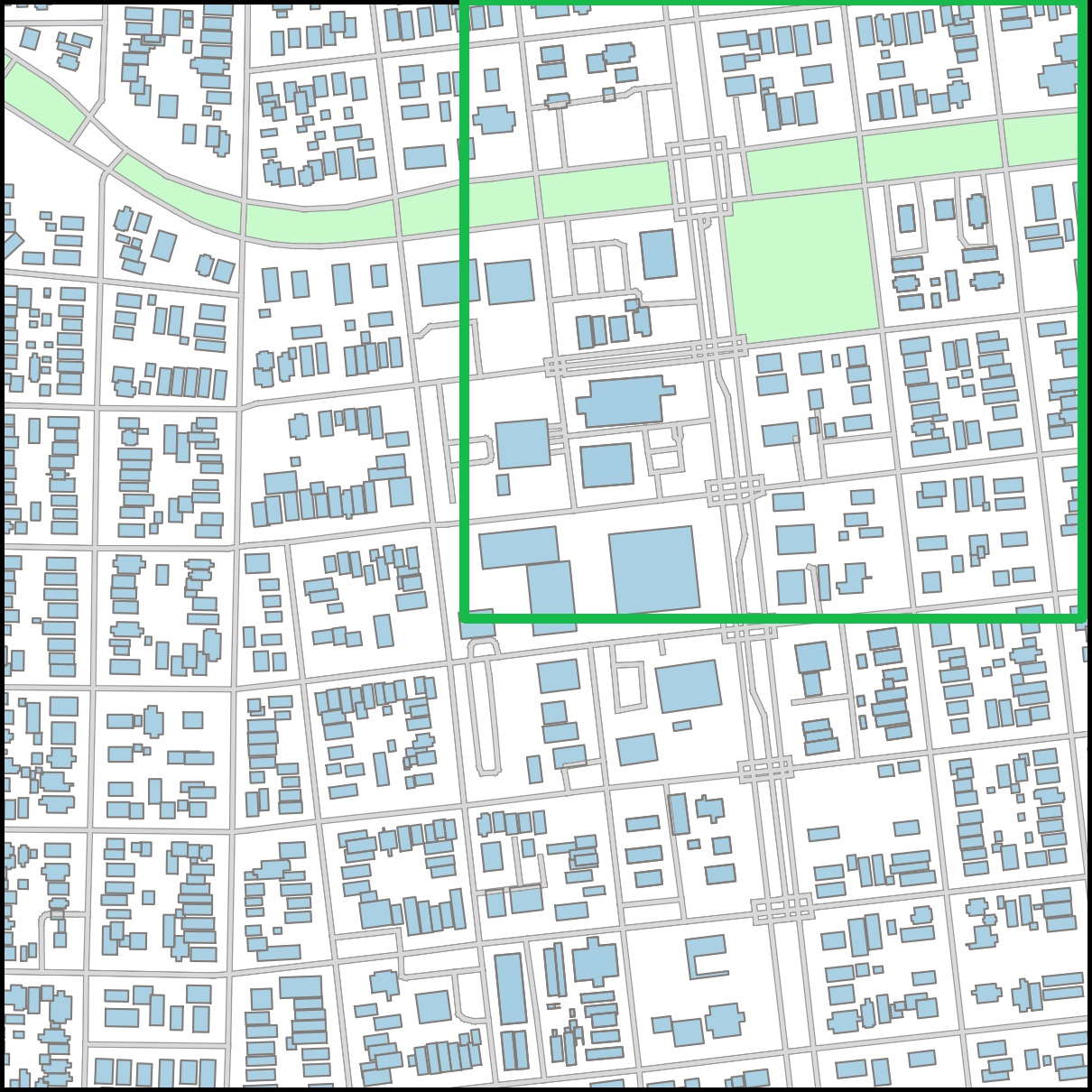} \\


   \includegraphics[width=0.19\textwidth]{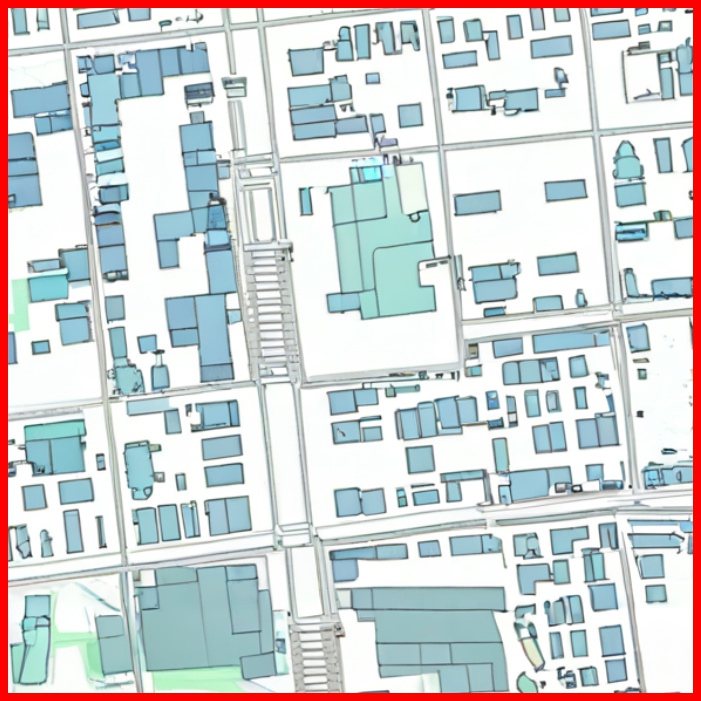} &
   \includegraphics[width=0.19\textwidth]{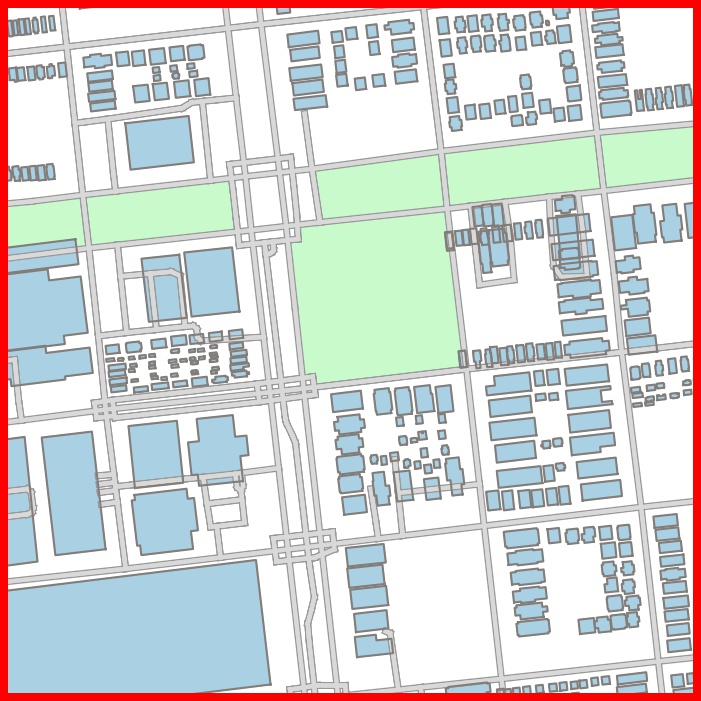} &
   \includegraphics[width=0.19\textwidth]{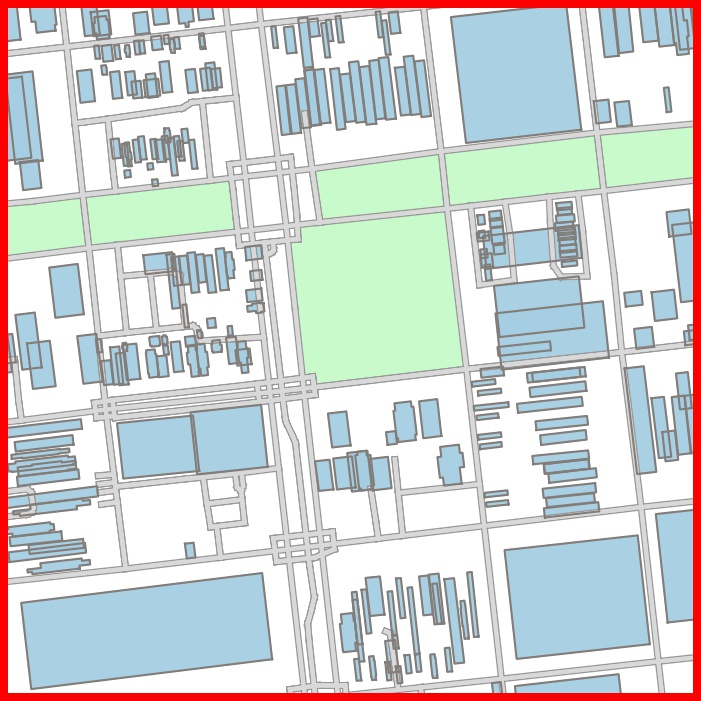} &
   \includegraphics[width=0.19\textwidth]{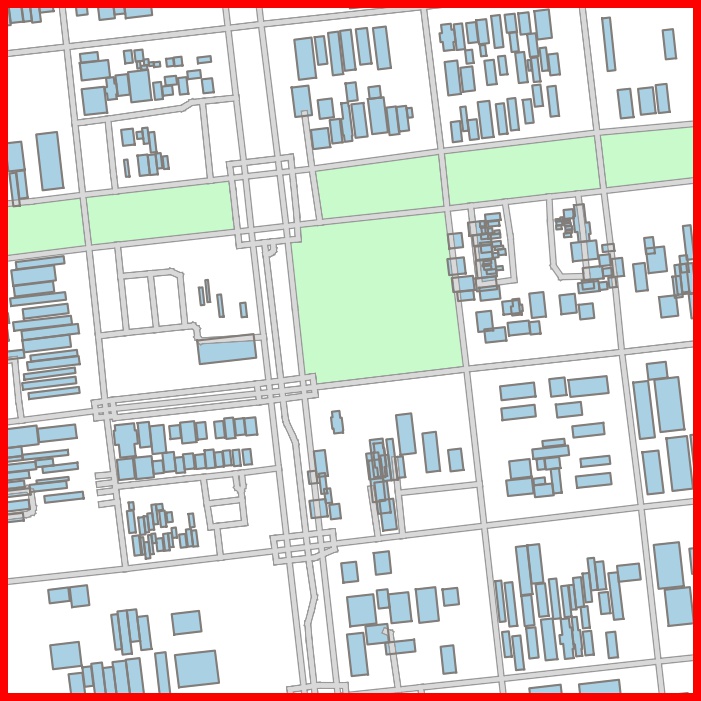} &
   \includegraphics[width=0.19\textwidth]{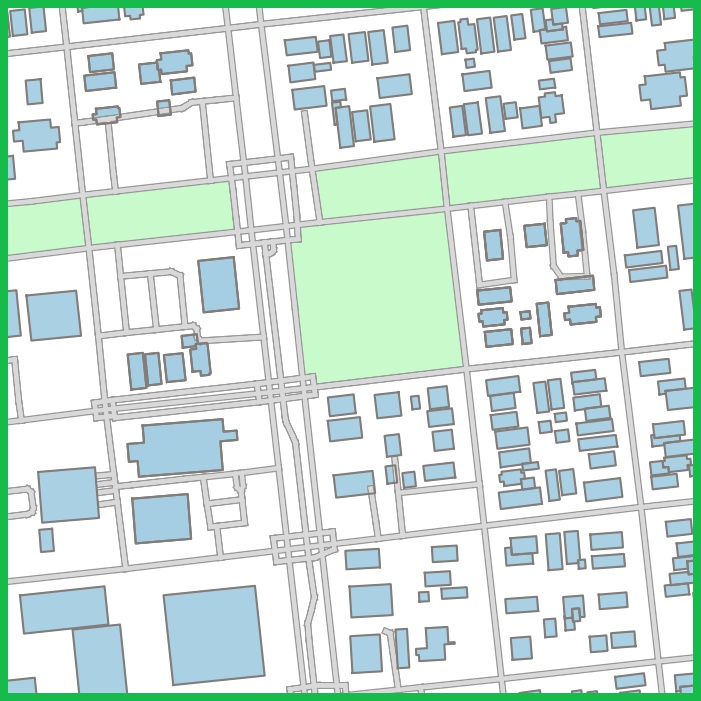} \\

    \includegraphics[width=0.19\textwidth]{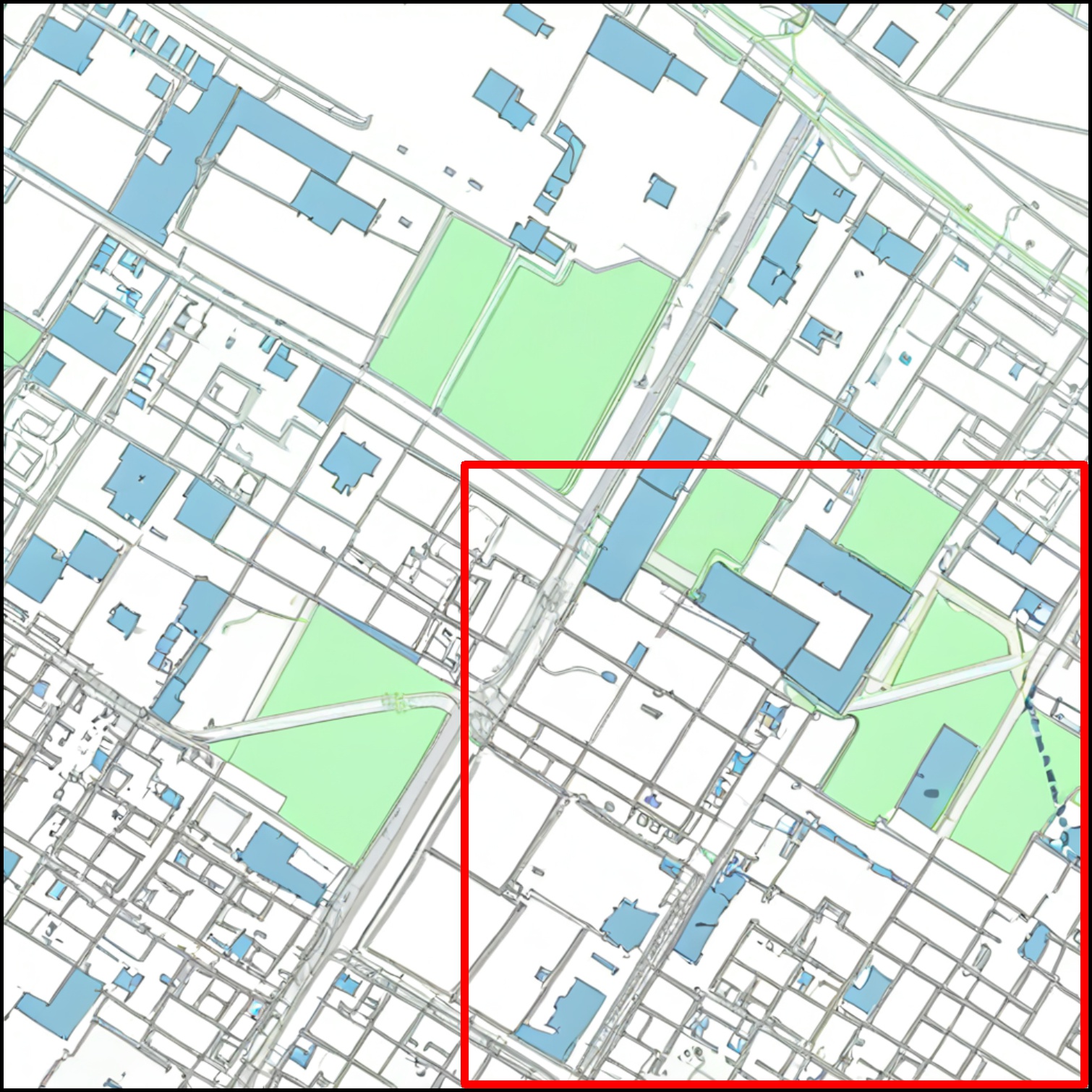} &
    \includegraphics[width=0.19\textwidth]{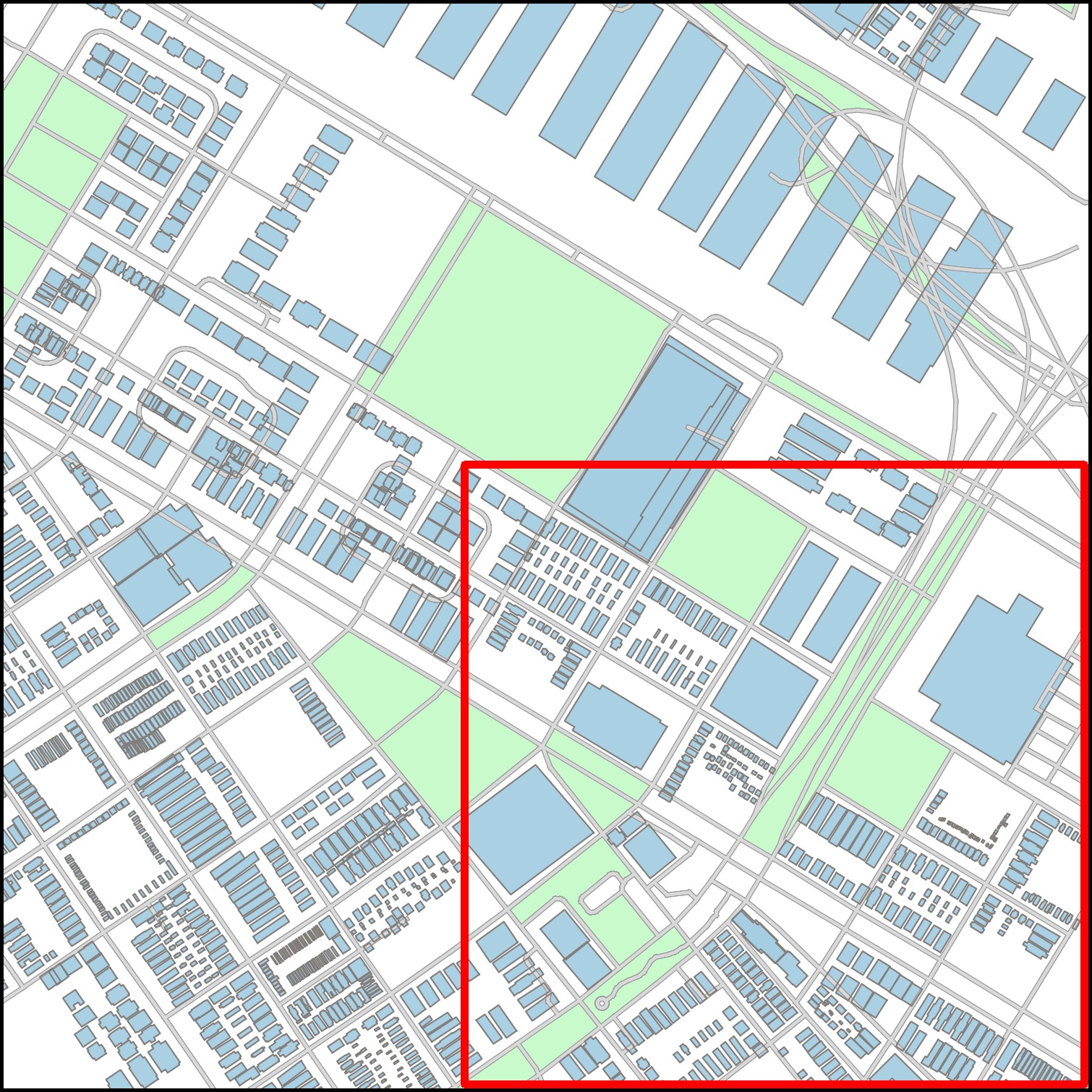} &
    \includegraphics[width=0.19\textwidth]{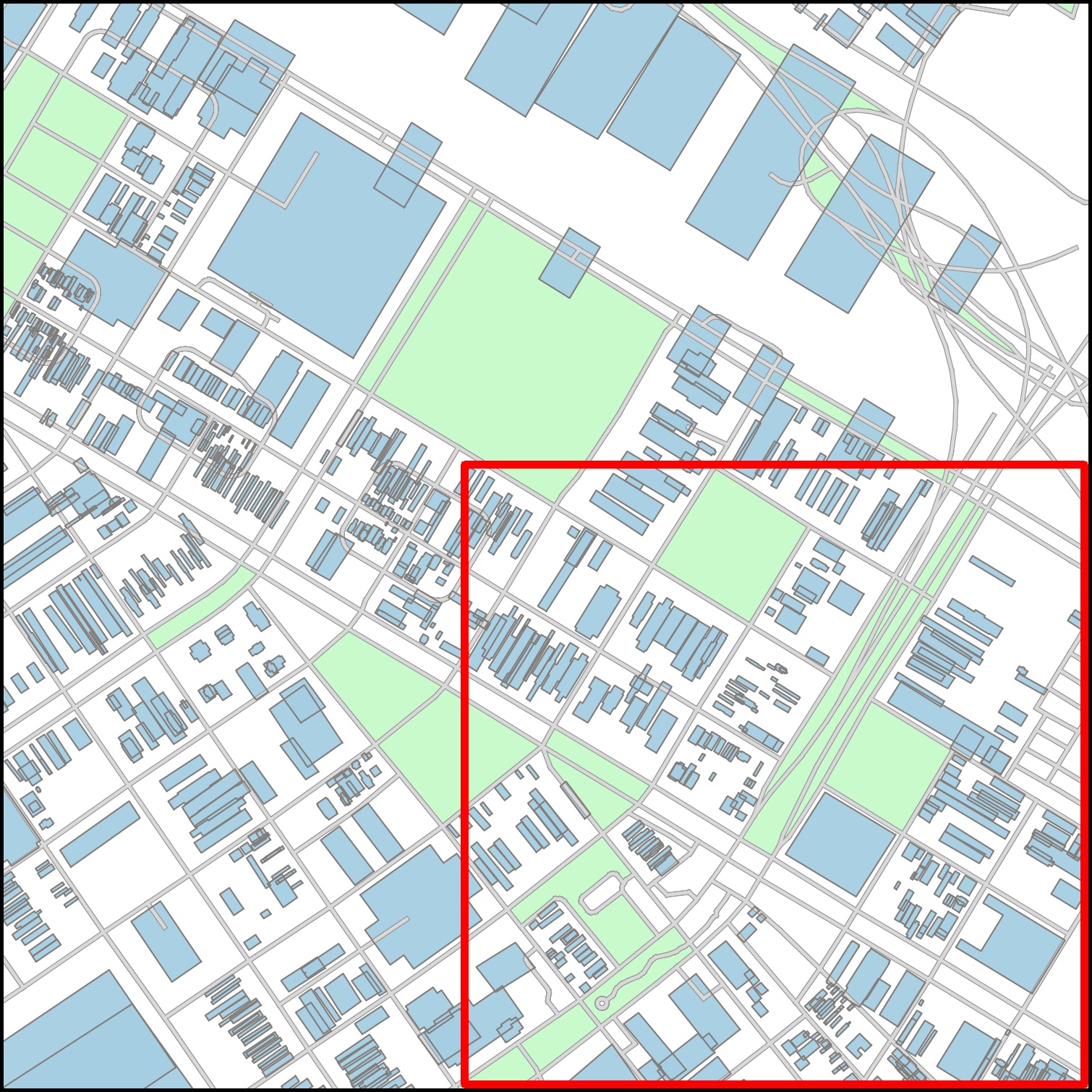} &
    \includegraphics[width=0.19\textwidth]{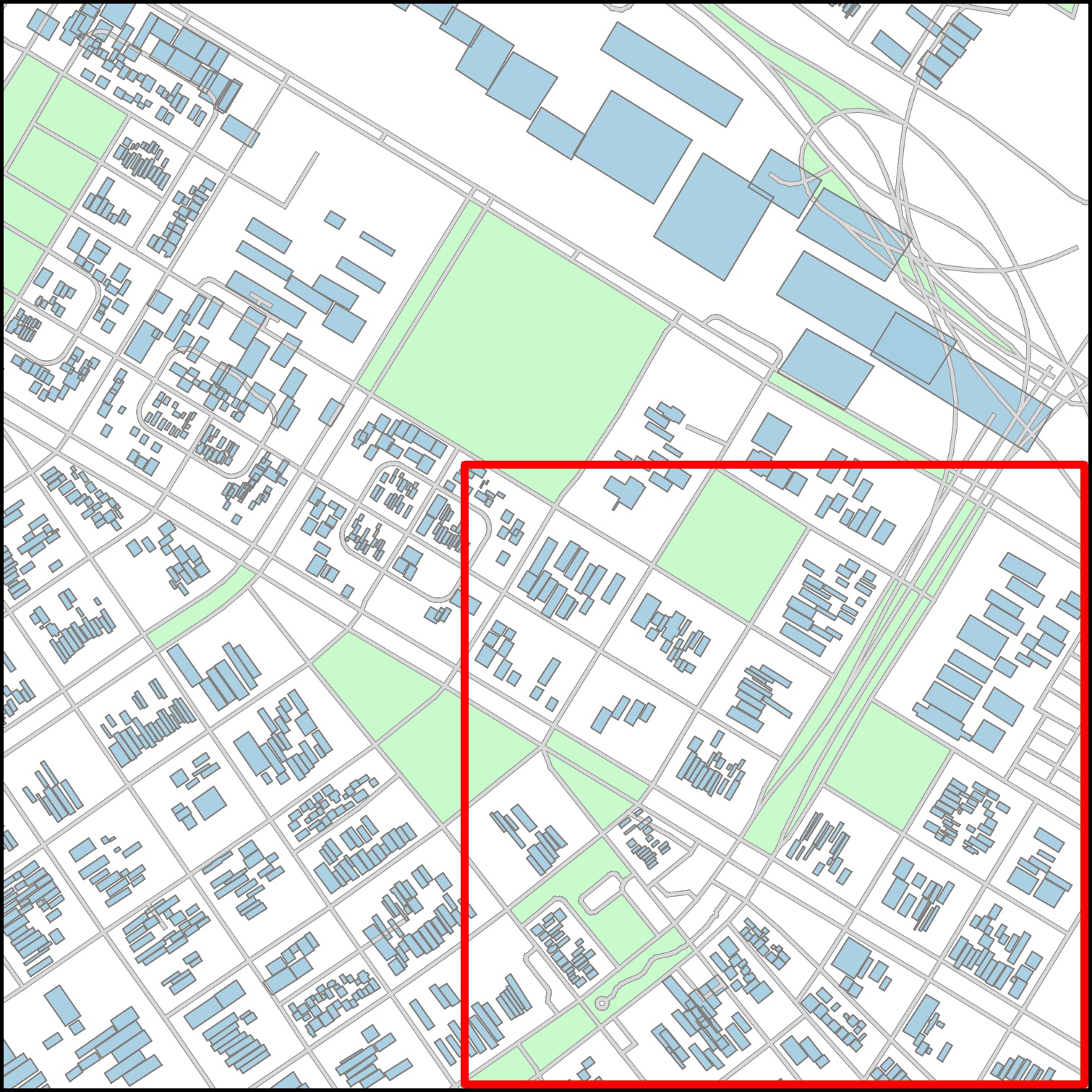} &
    \includegraphics[width=0.19\textwidth]{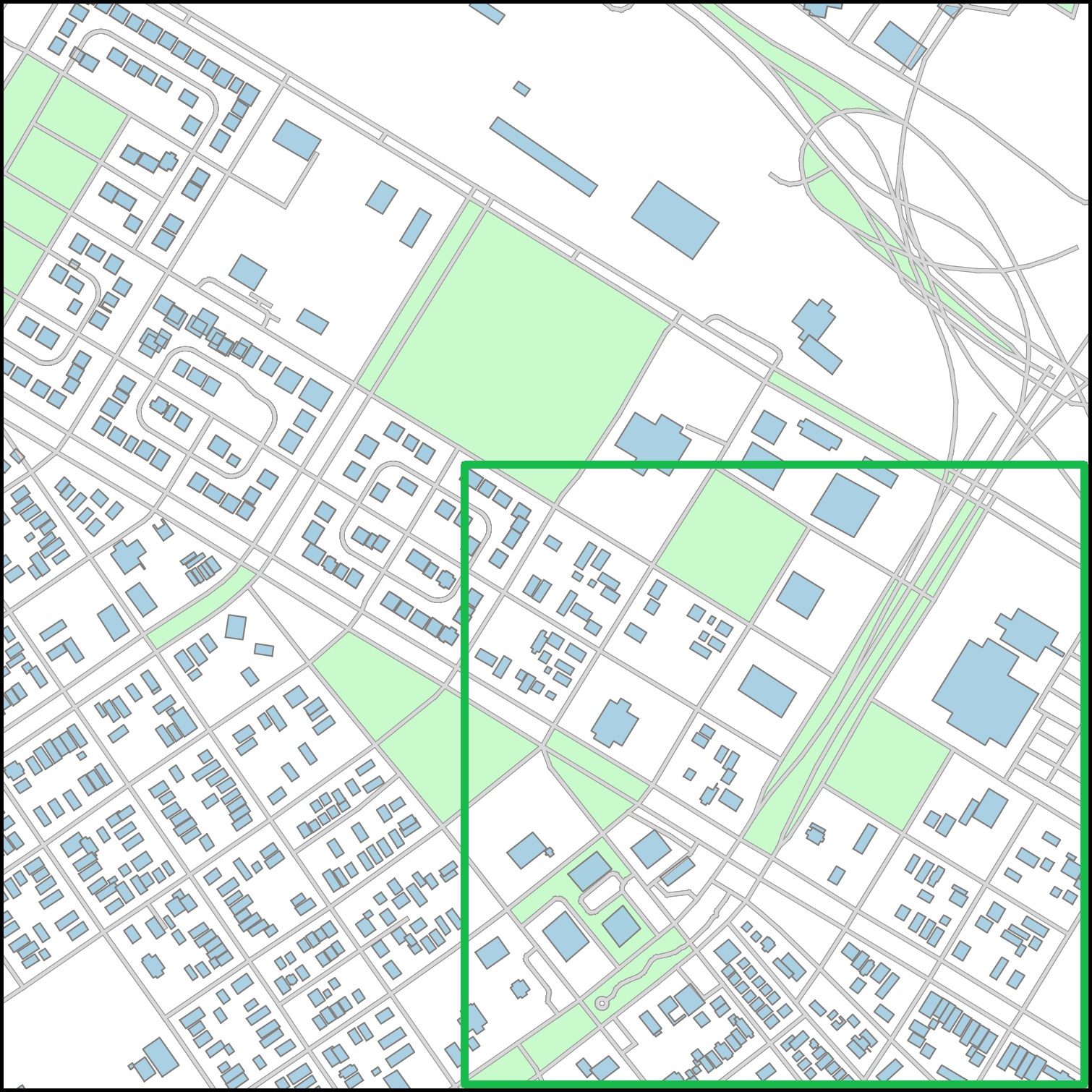} \\

   \includegraphics[width=0.19\textwidth]{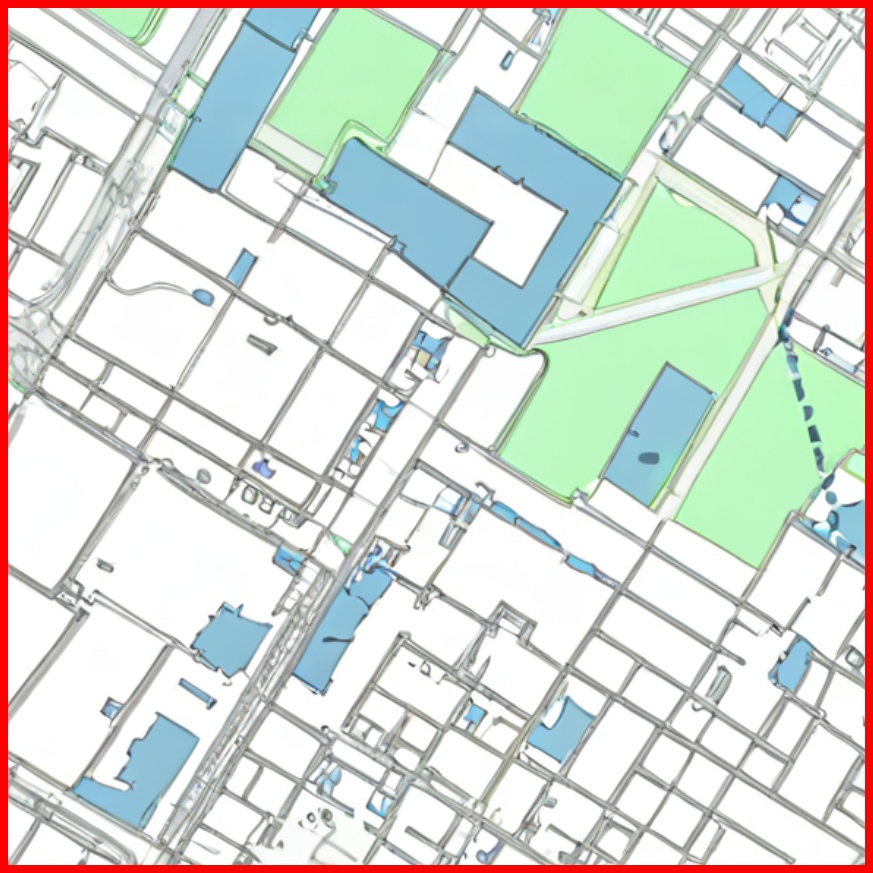} &
   \includegraphics[width=0.19\textwidth]{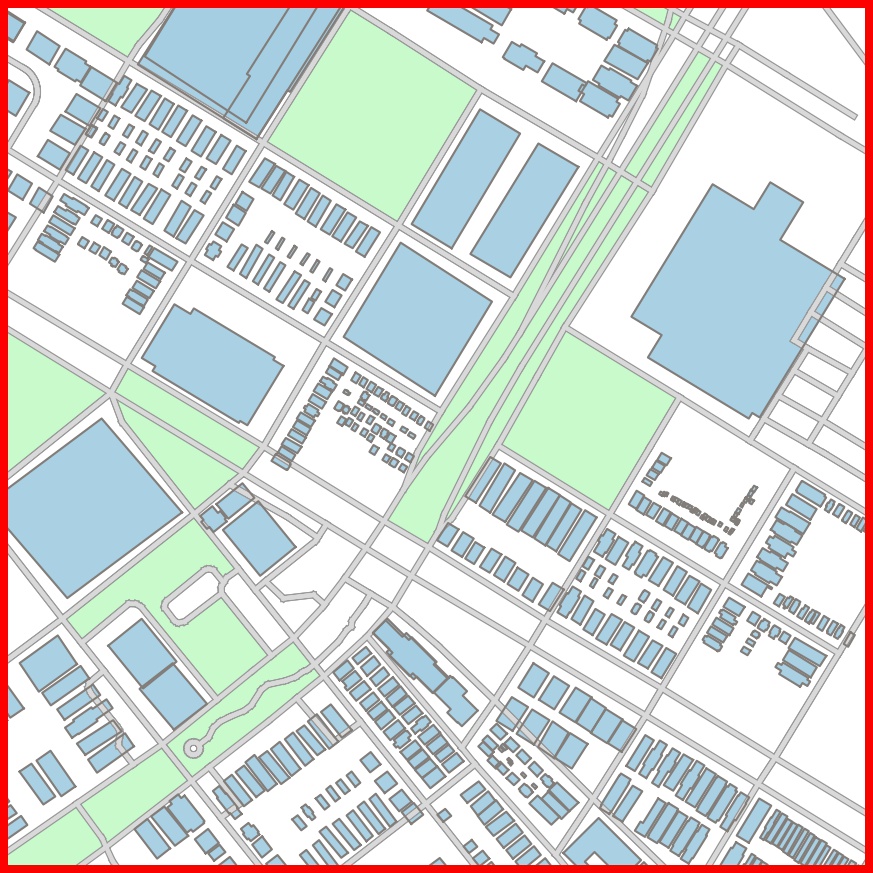} &
   \includegraphics[width=0.19\textwidth]{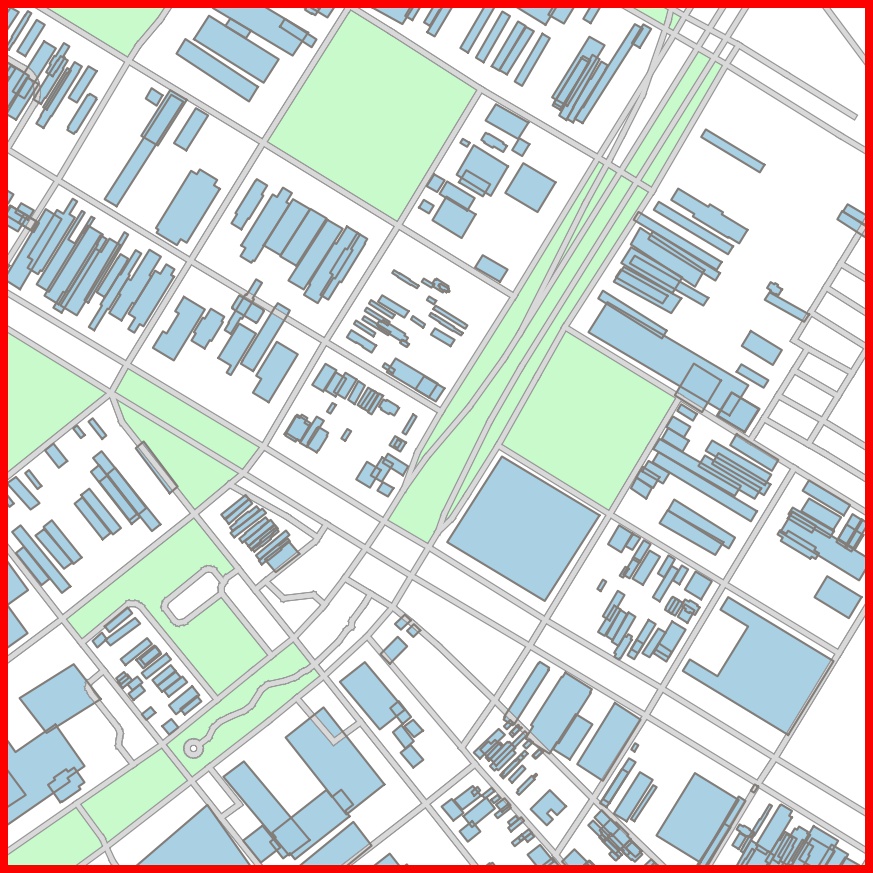} &
   \includegraphics[width=0.19\textwidth]{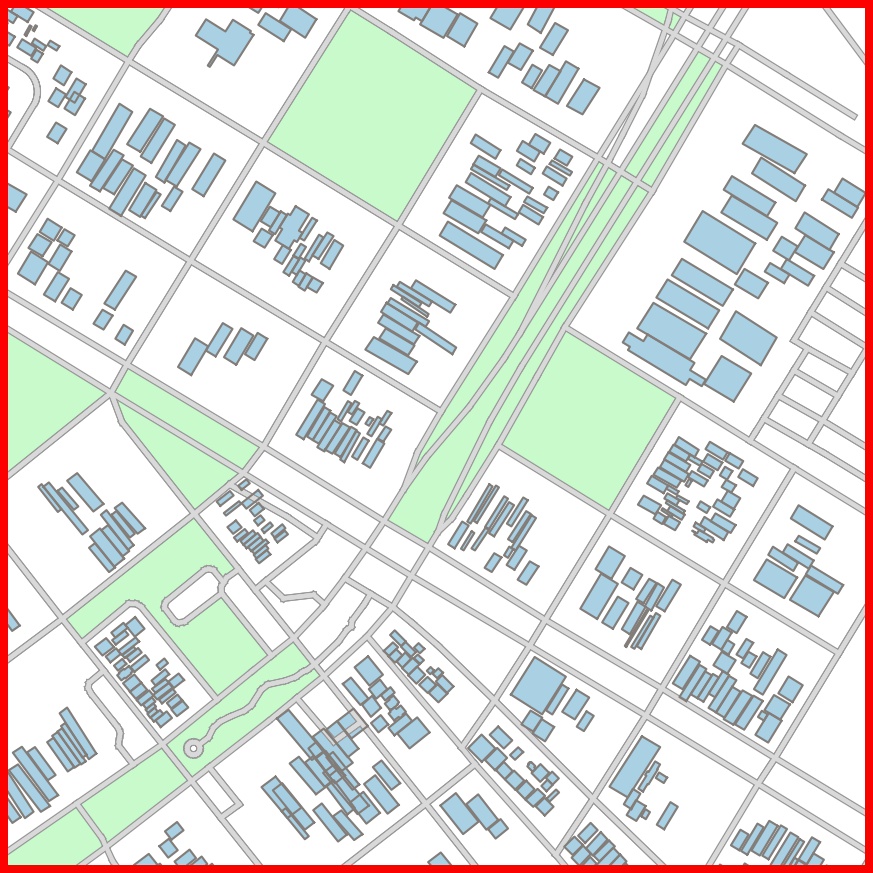} &
   \includegraphics[width=0.19\textwidth]{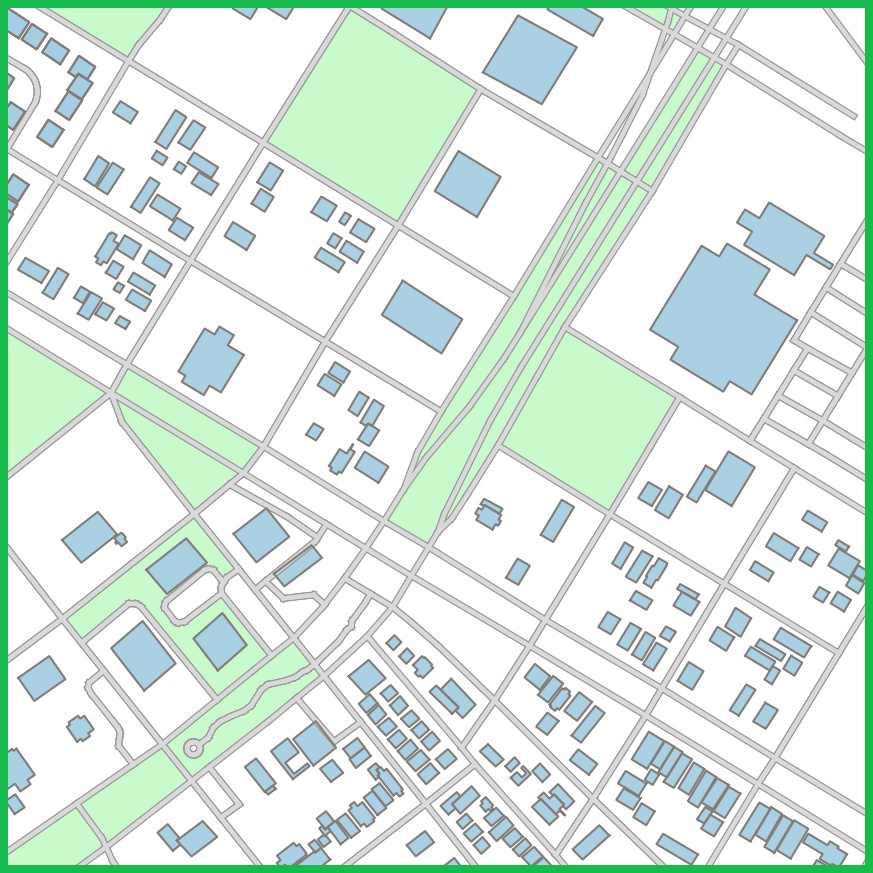} \\

    SDXL & VTN & LayoutDM& GlobalMapper & Ours
\end{tabular}%
    \caption{
    \textbf{Qualitative Comparisons.}  Given the same road network (except SDXL~\cite{podell2023sdxl}), all above methods generate urban layouts in only one pass without post-processing or human-in-the-loop refinement. The "even rows" are a zoom-in of the highlighted areas in the "odd rows". Our method generates a realistic distribution of urban layouts with plausible context-dependent behaviors (as indicated by $CTS$ score). VTN~\cite{arroyo2021variational} and LayoutDM~\cite{inoue2023layoutdm} show abnormal style shift among neighboring blocks, and no awareness of road networks. GlobalMapper~\cite{he2023globalmapper} generates over-similar communities. SDXL~\cite{podell2023sdxl} provides poor outpainting quality containing semantic errors and undesired overlaps. 
    }
  \label{fig:comparison}
\end{figure}

\subsection{Comparisons}
\label{sec:comparison}
We compare our method to several alternate city-scale layout generators (Tab.~\ref{tab:comparison}). Specifically, we randomly select 100 communities from among our test set, total 3700 city blocks. All methods generate the layout of the 100 communities in one pass, and without any post-processing or human-in-the-loop refinement. Implementations are downloaded from the author's repositories and trained with default parameter settings. For fairness, we convert our dataset to token sequences, graphs, or image patches (1024$\times$1024) as needed. VTN~\cite{arroyo2021variational},  LayoutDM~\cite{inoue2023layoutdm}, GlobalMapper~\cite{he2023globalmapper} are retrained for single block generation. SDXL~\cite{podell2023sdxl} is fine-tuned for outpainting by using the upper half to predict the lower half. 

\begin{table*}[tb]
  \caption{\textbf{Quantitative Comparisons.} All methods are compared to the same real urban layouts (except SDXL~\cite{podell2023sdxl} which cannot take-in a road network). Best values are in bold, second best values are underlined. Our method outperforms other existing methods in all but the overlap metric. See text for an explanation of the metrics.}
\footnotesize
\centering
\begin{tabular}{lccccccccc}
    \toprule

   Method & $CTS_{x \to 0}$ & WD-5D$\downarrow$ & WD-CO$\downarrow$ & Overlap$\downarrow$ & O-Blk$\downarrow$  &FID$\downarrow$ & KID$\downarrow$ & LPIPS$\downarrow$\\
    \midrule
   SDXL~\cite{podell2023sdxl}   & - & - & - & - & - & 120.24 & 0.079 & 0.48 \\
   VTN~\cite{arroyo2021variational} & -1.14 & 3.18 & 5.81 & \textbf{1.24} & 7.35 & 69.14 & 0.047 & \underline{0.32}\\
   LayoutDM~\cite{inoue2023layoutdm}  & -2.20 & \underline{2.92} & 12.50 & 4.56 & 1.72 & 66.77 & 0.040 & 0.39 \\
   GlobalMP~\cite{he2023globalmapper} & \underline{0.62} & 4.77 & \underline{4.14} & 2.52 & \underline{0.68} & \underline{49.55} & \underline{0.024} & 0.34\\
    \midrule
   \textbf{Ours} & \textbf{0.21}  &  \textbf{2.28}  & \textbf{1.91} & \underline{1.27} & \textbf{0.42} & \textbf{23.63} & \textbf{0.005} & \textbf{0.20}\\
    \bottomrule
\end{tabular}
\label{tab:comparison}
\end{table*}%

\subsubsection{Evaluation Metrics}
\label{subsec:evalutaion_metric}
We define a Context Score ($CTS$) to evaluate the context relationships among neighboring city blocks (i.e., a community). This metric is a numerical extension of LayoutSim~\cite{he2023globalmapper} and DocSim index~\cite{patil2020read} which evaluate the similarity between two city blocks regarding location, size, and category. Specifically, for each generated city block $b_{i}$, we calculate the average LayoutSim with its neighbors $N(i)$. Then, the local average of LayoutSim represents the context similarity among the community ($CT$). High $CT$ indicates identical blocks in a community and low $CT$ corresponds to high diversity in a community. $CTS$ is the subtraction of the real community's $CT$ from the generated $CT$:

\begin{equation}
\label{eqn:CTS1}
CT = \frac{1}{\|N(i)\|} \sum_{j}^{N(i)} LayoutSim(b_{i}, b_{j}), \quad N(i) = \{ j \mid (i, j) \in E \} 
\end{equation}

\begin{equation}
\label{eqn:CTS2}
CTS = CT_{gen} - CT_{real}
\end{equation}

 Given a real community, the metric indicates whether a generated community is over-diverse ($CTS<0$), or over-similar ($CTS>0$). A low value means more realism.

In our comparisons, we compute several metrics. One metric is the Wasserstein Distance between the distributions of generated communities and real communities in terms of 2D position (x, y) and 2.5D geometry (length, width, height), and building count (CO) -- we name these WD-5D and WD-CO. We define two quality metrics as well: Overlap indicates the percentage by which two building overlaps, and O-Blk represents the percentage of building area that sticks outside of its city block. We also implement several perceptual metrics: FID~\cite{heusel2017gans}, KID~\cite{binkowski2018demystifying}, and LPIPS~\cite{zhang2018unreasonable} to evaluate the visual quality and realism of generated layouts. For all the above metrics, lower value indicates better quality.

\subsubsection{Quantitative Results}
As Tab.~\ref{tab:comparison} shows, our approach performs better than existing approaches in all but one metric. Note that the poor quality of SDXL results makes it hard to decipher layouts. We only compare to it by pixel-based perceptual metrics.

\subsubsection{Qualitative Results}
We show in Fig.~\ref{fig:comparison} qualitative results for several exemplary areas (more examples are in Supplemental Material). In general, our method shows good contextual harmonization and has awareness of arbitrary road networks. It generates a realistic distribution of urban layouts with plausible context dependent behaviors. Other methods show improper patterns (e.g., overly diverse or similar) and potentially containing semantic errors or undesired overlaps.

\subsection{Ablation Studies}
\label{sec:ablation}

\begin{table}[tb]
\footnotesize
\centering
\caption{\textbf{Block Quantization Ablations.} We report metrics (all in $\%$) among all alternative ablations of our Block-level graph-based Variantional AutoEncoder (BVAE). Using GAT-backbone with post-trained dimensional quantization with $L=20$ shows the best overall compromise. The best values are in boldface. }
\label{tab:blk_quant_ablation}

\begin{tabular}{lcccccc}
    \toprule
   \textbf{Encode Model} & Overlap$\downarrow$ & Out-Blk$\downarrow$ & Pos-E.$\downarrow$ & Geom-E.$\downarrow$ & Ct-E.$\downarrow$ & Cov-E.$\downarrow$\\
    \midrule  
    LayoutVAE~\cite{Jyothi19} & 13.84 & 11.15 & 15.09 & 37.35 & - & 24.34 \\
    VTN~\cite{arroyo2021variational} & 2.70 & 4.70 & 6.01 & 22.86 & 18.58 & 1.97\\
    BlockPlanner~\cite{xu2021} & 2.93 & 1.30 & 4.78 & 12.06 & 3.63 & 5.66 \\
    GlobalMapper~\cite{he2023globalmapper} & 1.04 & \underline{1.20} & \underline{3.25} & 2.83 & 0.08 & 0.46\\
   BVAE (SAGE~\cite{hamilton2017inductive}) &  1.33 & 1.70 & 5.01 & 3.25 & 0.10 & 0.35\\
   BVAE (GCN~\cite{kipf2016semi})     & \underline{0.45} & 1.21 & 4.25 & \underline{2.74} & \underline{0.04} & \underline{0.28}\\
    \textbf{BVAE*} (GAT~\cite{velivckovic2017graph}) & \textbf{0.17} & \textbf{0.28} & \textbf{1.00} & \textbf{0.71} & \textbf{0.006} & \textbf{0.04}\\
    \midrule
    \\
    \midrule
    \textbf{Quantization} & Overlap$\downarrow$ & Out-Blk$\downarrow$ & Pos-E.$\downarrow$ & Geom-E.$\downarrow$ & Ct-E.$\downarrow$ & Cov-E.$\downarrow$\\
    \midrule  
   Trainable-VQ   & 3.02 & 0.55 & 3.75 & 5.78 & 0.07 & 0.60 \\ 
   KMeans-Q & 2.90 & \underline{0.02} & 20.98 & 17.85 & 50.01 & 16.23\\
   DIM-Q ($L=5$) &  0.28 & 0.62 & 7.71 & 16.20 & 25.0 & 14.68  \\
   DIM-Q ($L=10$) & 0.56 & 0.28 & 5.73 & 4.48 & 8.33 & 4.58    \\
   DIM-Q ($L=30$) & \underline{0.17} & \textbf{0.01} & \textbf{2.68} & \textbf{0.78} & \textbf{0.005} & \underline{0.15}  \\    
    \textbf{DIM-Q* ($L=20$)}  & \textbf{0.11} & \textbf{0.01} & \underline{3.50} & \underline{0.89} & \underline{0.006} & \textbf{0.12}   \\
    \bottomrule

\end{tabular}
\end{table}%

\subsubsection{Block Quantization}
 We ablate block quantization in order to understand and maximize the impact of changing its parameters and design strategy. First, we report in Tab.~\ref{tab:blk_quant_ablation} (Encode Model Table) the reconstruction loss using different block-level variational autoencoders. We evaluate using various geometric metrics: Overlap (percent of building-to-building overlap), Out-Blk (percent of buildings that stick outside of its city block), Pos-E (positional error as a percentage of block diagonal), Geom-E (2.5D building geometry error as percentage of block scale), Ct-E (percentage of building count error), and Cov-E (total building coverage error). They clearly show our approach is the best overall. Thus we choose our BVAE based on Graph Attention Networks (GAT)~\cite{velivckovic2017graph}.

Then, in Tab.~\ref{tab:blk_quant_ablation} (Quantization Table), we ablate various codebook approaches and quantitization levels for block-level encoding. "Trainable-VQ" leverages trainable vector quantization codebooks introduced by VQVAE~\cite{van2017neural}. We also provide an alternative non-training based quantization method using KMeans ("KMeans-Q") to cluster BVAE latent space values to a reasonable number of categories in order to handle diverse urban layouts (e.g., 100 classes). The "DIM-Q" rows imply quantizing each of the 512 dimensions of the block-level code into a codebook $C=\{1, 2, ..., L\}$ as described in Sec.~\ref{sec:canonical_repr}. This per-dimension quantization performs clearly the best. Specifically, setting $L$ to infinity produces a lossless representation of BVAE latent vectors, but also causes the curse of dimensionality. We discover that $L=20$ per dimension is the best overall compromise balancing compactness and representation ability.

\begin{table*}[tb]
  \caption{\textbf{City-Scale GMAE Ablations. } We ablate backbone message passing layers, maximum messaging passing hops ($D$: depth of encoder), masking strategy during training, and scheduling functions of iterative generation ($T$ indicates total iterations). The overall best setting is marked as $^*$.}
\scriptsize
\centering
\begin{tabular}{lccccccccc}
    \toprule
   Ablations & Context$\downarrow$ & WD-3D$\downarrow$ & WD-N$\downarrow$ & Overlap$\downarrow$ & Out-Blk$\downarrow$  & FID$\downarrow$ & KID$\downarrow$ & LPIPS$\downarrow$\\
    \midrule
   GMAE (GCN~\cite{kipf2016semi}) & -0.30 & 2.74 & 2.91 & \textbf{0.26} & 0.36 & 52.95 & 0.025 & 0.32 \\
   GMAE (GTrans~\cite{shi2020masked}) & 0.74 & 3.02 & 3.27 & 1.96 & 1.07 & 43.60 & 0.018 & 0.29 \\
   GMAE (SAGE~\cite{hamilton2017inductive}) & 0.97 & 2.98 & 2.36 & 1.75 & 2.60 & 43.15 & 0.017 & 0.29\\
    \midrule
   GMAE (GAT, Dec) & 0.94 & 2.45 & 2.96 & 2.13 & 1.10 & 48.24 & 0.019 & 0.30\\
   GMAE (GAT, $D=1$) & 1.25 & 3.03 & 2.64 & 1.73 & 0.44 & 43.75 & 0.018 & 0.29\\
   GMAE (GAT, $D=2$) & 0.33 & 2.66 & \underline{2.01} & 1.31 & 0.57 & 34.82 & 0.014 & 0.27\\
   GMAE (GAT, $D=4$) & \underline{0.24} & \textbf{1.92} & 2.31 & \underline{0.63} & \textbf{0.28} & \underline{24.59} & \underline{0.009} & \underline{0.22}\\
    \midrule
   GMAE ($Mask\in[0.05, 1]$) & 1.30 & 2.70 & 3.62 & 1.70 & 2.45 & 43.14 & 0.018 & 0.29\\
   GMAE ($Mask\in[0.05, 0.5]$) & 1.10 & 3.07 & 3.68 & 1.68 & 1.30 & 46.85 & 0.020 & 0.30\\
   GMAE ($Mask = 0.15$) & 1.59 & 4.91 & 3.41 & 1.83 & 1.20 & 49.50 & 0.022 & 0.31\\
    \midrule
   GMAE (Linear, $T=12$) & 0.63 & 2.83 & 3.79 & 1.27 & 0.93 & 37.42 & 0.015 & 0.28\\
   GMAE (Log, $T=12$) & -1.31 & 2.77 & 7.27 & 7.41 & 3.14 & 63.25 & 0.038 & 0.33\\    
   GMAE (Cosine, $T=1$) & -1.67 & 7.44 & 5.64 & 3.56 & 4.01 & 76.12 & 0.051 & 0.35\\    
   GMAE (Cosine, $T=20$) & 0.32 & 2.92 & 3.98 & 1.68 & \underline{0.31} & 32.06 & 0.013 & 0.24\\    
    \midrule
   \textbf{\tiny *(GAT-D3, [0.5,1], Cos-T12)} & \textbf{0.21} &  \underline{2.28}  & \textbf{1.91} & 1.27 & 0.42 & \textbf{23.63} & \textbf{0.005} & \textbf{0.20}\\
    \bottomrule
\end{tabular}
\label{tab:comparison2}
\end{table*}%

\subsubsection{City-Scale Graph Masked Autoencoder.}
In Tab.~\ref{tab:comparison2}, We ablate various model parameters, together with significant training and inference strategies for our GMAE. Each alternatives is evaluated by the same set of metrics in Sec.~\ref{sec:comparison}.

By varying the message passing layers, we find GAT~\cite{velivckovic2017graph} provides the best performance. GCN~\cite{kipf2016semi} produces diverse and unrealistic urban layouts, while transformer-based~\cite{shi2020masked}, and GraphSAGE~\cite{hamilton2017inductive} backbones generate blocks that are too similar. The depth of the stacked layers in the encoder (D) dictates the maximum message passing hops from each node and thus the span of context behaviors captured for realistic generation. We find that 3 hops is promising and larger contexts cause over-similarity and unnecessary computing burden. We also observed that using stacked GAT layers for decoding with remasking tricks, as described by~\cite{hou2022graphmae,hou2023graphmae2}, didn't produce a net benefit towards city generation. So we choose MLP as the node feature decoder for both its performance and efficiency.

The masking strategy during training is crucial for GMAE performance. We find that dynamic and large masking ratios are beneficial, while a fixed small ratio (e.g., 0.15 as in~\cite{devlin2018bert}) leads to over similarity. This behavior is also observed by Li et al.~\cite{li2023mage}. We adopt a similar masking strategy by sampling a Gaussian distribution $N(0.55, 0.25)$, truncated by $[0.5, 1.0]$, for each training iteration.

Scheduling for iterative generation is also critical to capture representative city blocks (Sec.~\ref{sec:iterative_sample}). We evaluate several functions $\beta(t)$ (e.g., linear, logarithmic, or cosine-based) that provide the scheduled acceptance ratio, and evaluate alternative total iteration counts $T$. Rapid iteration leads to over-diversity and long iteration results in over-similarity. Logarithmic functions with reverse pattern produces the worst realism. Metrics show that cosine-based function, which is conservative initially and generous in later iterations, outperforms others.

In summary, this study determines the best configuration for our GMAE as: GAT encoder with $D=3$ and MLP decoder, training mask ratio $m\in[0.5, 1.0]$, and priority-based scheduling function $\beta(t)=1-cos(t/T)$ with $T=12$.

\subsection{Limitations}
Our method faces limitations where shapes of city blocks or buildings are not simple polygons (e.g. the building with a garden inside). It also struggles to handle very irregular and concave shaped city blocks (e.g., cul-de-sac's). Also, road-network to contextual structure relations are not explicitly considered.


\subsection{Additional Results}
We place several other interesting results in Supplementary Materials.

\begin{itemize}
    \item[$\bullet$] \textbf{Controllable Generation of Entire City.} Our method enables realistic and efficient city-scale controllable generation given any percentage of priors. We show several large examples spanning a subset of the many cities.
    \item[$\bullet$] \textbf{Socio-Economic Prediction.} Trained GMAE may act as a powerful encoder for downstream tasks. For example, the encoded GMAE latent vectors of urban layouts can indicate social-economic metrics for each city block.
    \item[$\bullet$] \textbf{Semantic Manipulations.} Our method enables semantic manipulations over various building layout styles across different cities.
\end{itemize}

\section{Conclusions and Future Works}
Large-scale urban layout generation has seen significant advancements, spurred by the intersection of computer vision, urban planning, and related disciplines. Previous methodologies often neglect the context-sensitive nature of urban layouts, failing to capture the multi-layer semantics crucial for realism and plausibility. Our proposed Graph-based Masked AutoEncoder (GMAE) addresses these limitations by leveraging a graph-based representation, acknowledging context-sensitivity, and prioritizing the generation process. By training on a diverse dataset of 330 cities, we demonstrate the efficacy of our approach in generating large-scale 2.5D layouts with realism and semantic consistency. Moreover, our method outperforms existing techniques in various metrics, marking a significant advancement in the realm of urban layout generation. 

Moving forward, our approach paves the way for more accurate and efficient urban simulation, digital twin creation, and game/content design. As future work we intend to incorporate our method for large-scale photo-realistic multi-view scene synthesis, and city-scale 3D modeling. Those potentials will contribute to synthetic data generation in autonomous driving and world model training.

\section*{Acknowledgements}

This project was funded in part by NSF Grant \#2107096 and NSF Grant
\#1835739.

%
%
\bibliographystyle{splncs04}
\bibliography{main}

\pagebreak
\begin{center}
\noindent \textbf{\Large Supplementary Materials of: COHO: Context-Sensitive City-Scale Hierarchical Urban Layout Generation}

\setcounter{section}{0}

\end{center}
Below is a summary of the contents in each section of this supplemental material:
\begin{itemize}
\item[$\bullet$] Sec.~\ref{supp_sec:supp_dataset}: Details of our dataset collection method.
\item[$\bullet$] Sec.~\ref{supp_sec:supp_quantization}: Details of building layout quantization, including graph-based Block-level Variational AutoEncoder (BVAE), and different quantization approaches. We also provide additional evaluation in real units (e.g. $m$, $m^2$).
\item[$\bullet$] Sec.~\ref{supp_sec:big_examples}: City-scale controllable generation examples for 17 cities. And some complementary comments on other alternative approaches.
\item[$\bullet$] Sec.~\ref{supp_sec:semantic_manipulation}: Semantic manipulations between different building layout styles. 
\item[$\bullet$] Sec.~\ref{supp_sec:socio_eco}: GMAE-based Socio-Economic Metric Prediction
\item[$\bullet$] Sec.~\ref{supp_sec:add_ablation}: Estimation of running time and model parameter sizes.

\end{itemize}

\section{Dataset Collection}
\label{supp_sec:supp_dataset}
\subsubsection{Data Resources.} We selected all US cities with more than 100K population (i.e., 330 cities) as our building layout dataset. The city list is based on~\cite{citylist}. For each city, we define a rectangular bounding box containing most of the metropolitan area. This geo-registered bounding box is used to extract data from OpenStreetMap (OSM)~\cite{OpenStreetMap}, Microsoft Building Footprints (MSF)~\cite{heris2020rasterized}, and Topologically Integrated Geographic Encoding and Referencing (TIGER) dataset~\cite{tiger}. The TIGER dataset provides the city block contour and corresponding social economic metrics (e.g. population, income level, etc.). 

MSF and OSM provide the building layout polygons with corresponding height values. We define heuristics to composite both data resources. Specifically, MSF is the result of deep-learning-based building segmentation from high-resolution satellite images (~$0.5m$). However, fine details and closely located buildings are challenging for that methodology. Such a situation is very common in big cities like New York (i.e., a big city block appears to have one large building but in reality there are many adjacent buildings). In contrast, the building layouts from OSM are manually crafted or adopted from governmental/open-source datasets. Buildings are vector-based polygons regardless of adjacency. Unfortunately, the data coverage of OSM, especially for building height values, varies significnantly, while satellite-resourced MSF overcomes this shortage. We merge the information from the two sources as described in the following section.

\subsubsection{Data Composition.} For each single city block, we evaluate and composite the building layouts extracted from both MSF and OSM by several rules: (1) we keep buildings from both resources that only appear once with no mutual overlapping; (2) for the regions that building layouts from both resources are highly overlapped, we keep the ones with more building numbers; and (3) we mainly take building height values from MSF. If height values are only available for some buildings in a city block, then we take the average height augmented with small random offsets and assign those heights to the rest of the buildings. Further, the connectivity between city blocks is captured as a graph adjacency matrix for our graph canonical representation as specified in the main paper. Each city is encoded as a single graph. Thus our dataset contains 330 graph structures. It includes 833,473 city blocks and 17,663,607 buildings, all with building heights assigned. As we claimed in the main paper, we will release the dataset upon the acceptance of our paper.

\section{Building Layout Quantization}
\label{supp_sec:supp_quantization}
The main goal of building layout quantization is to define a compact and scalable representation (e.g., a small codebook) that keeps as much of the original building layout features as possible. In the following, we report our various experiments until arriving at the solution we ultimately used.

\subsection{Block-level Variational AutoEncoder (BVAE)}
To represent individual city blocks, we use a variational autoencoder. In this, all building layout features (e.g., building size, height, position) within a city block are represented as a canonical graph. This graph-representation is utilized in self-supervised training with both reconstruction loss and KL divergence loss terms. Our block representation is based on the canonical spatial transformation and graph-based representation described in GlobalMapper~\cite{he2023globalmapper} (official repository: \href{https://github.com/Arking1995/GlobalMapper}{GlobalMapper}). 

Our BVAE uses GAT~\cite{velickovic2017graph} as the backbone. The encoder stacks three GAT layers and the same structure for the decoder. Specifically, we select the multi-head number for GAT as 12, the internal feature dimension of GAT layers as 256, and the bottleneck latent dimension as 512. We found the aforementioned hyper-parameters are key to the reconstruction performance of BVAE. Reducing these parameter values hurts performance, and further increasing has no net benefit. Batch normalization is added to each layer for stable convergence. Our code will be released upon the acceptance of our paper.

\subsection{Quantization Approaches}
\subsubsection{Trainable Vector Quantization.} We experimented with the trainable vector quantization method described in VQGAN~\cite{esser2021taming} by adapting their official implementation  \href{https://github.com/CompVis/taming-transformers}{VQGAN} to our BVAE. After performing a round of hyperparameter tuning experiments, we did not converge to a trained result that adequately reconstructs the original building layouts. In particular, the network struggles to keep stability between the BVAE latent space and the trained codebook -- e.g., when inspecting latent space values near a codebook entry, unstable block structures are created. A predefined codebook size may not be able to adapt to the heterogeneity of many city block and graph feature values which, unlike pixel values in $\in$[0, 255], are not constrained by a given range. Moreover, our hyperparameter search indicates performance is even harder to be improved by increasing codebook size and feature dimensions.

\subsubsection{Latent Vector Clustering.} Given the trained BVAE encoder, one straightforward methodology is to use its encoded latent vectors as building layout features $Q$ for our subsequent GMAE training. We tried the simple approach of directly using the latent veectors but GMAE training convergence was not possible. Instead, we used K-means clustering to group latent space vectors into a set of classes. Thus, each city block will essentially be classified into one of a set of categorical labels. We tested using from 5 to 1000 clusters. However, this approach brought unacceptable reconstruction loss likely due to the heterogeneity of city blocks. We list the best performance of both K-means quantization, and trainable vector quantization in the Tab.2 (Quantization Table) in the main paper. Both are not comparable to the dimensional quantization as described in the main paper Sec. 3.1.

\subsubsection{Dimensional Quantization.} The key idea for our dimensional quantization is to use a small codebook (e.g. $L=20$) for each dimension of the latent space vector. Specifically, the BVAE Encoder outputs a 512-d vector of $\mu$ and another 512-d $\sigma$ for variational sampling of the network bottleneck. We find that the basic styles (e.g., building number, positions) of building layouts decoded by trained BVAE Decoder is generally deterministic to $\mu$, while $\sigma$ provides reasonable randomness. Thus we  directly sample from the full distribution of $\sigma$ from training dataset, and only consider the prediction of the $\mu$ (i.e., the 512 dimensions of $q_{i}$). A similar trade-off is also introduced in diffusion model training (e.g.,~\cite{ho2020denoising}). 

Our dimensional quantization is driven by the observation that the  distribution in each dimension of $\mu$ is nearly normal. Moreover, the trained BVAE Decoder is robust to values near the normal distribution's mean. This enables quantization of each dimension with good stability and a low reconstruction loss (e.g., a slight change of values does not affect the decoded building layouts significantly). We choose $L=20$ based on the ablations in Tab. 2 (Quantization Table) in the main paper.

\subsection{Additional Real-World Units Evaluation}

We evaluate the 2.5D building geometry as per-block percentage of "GEOM-E", and "Pos-E" in the Tab. 2 of the main paper. Here we provide a conversion to real-world units for better understanding. The row "BVAE$^*$" in Tab.2 (Encode Model Table) writes 1.00\% Pos-E, and 0.71\% Geom-E. This corresponds to each reconstructed building having an average height error of $0.10m$, size error of $16.97m^2$, and position error of $4.07m$. The row "DIM-Q$^*$ ($L=20$)" in Tab. 2 writes 3.50\% Pos-E, and 0.89\% Geom-E. This corresponds to each single building having a height error of $0.27m$, size error of $27.52m^2$, and position error of $8.07m$.

The increasing errors of position and 2.5D geometry are mainly due to the numerical quantization in each $\mu$ dimension and the randomly sampled $\sigma$, as described in above sections.

\section{City-Scale Controllable Generation}
\label{supp_sec:big_examples}

\begin{figure*}
\centering

\frame{\includegraphics[width=0.95\textwidth]{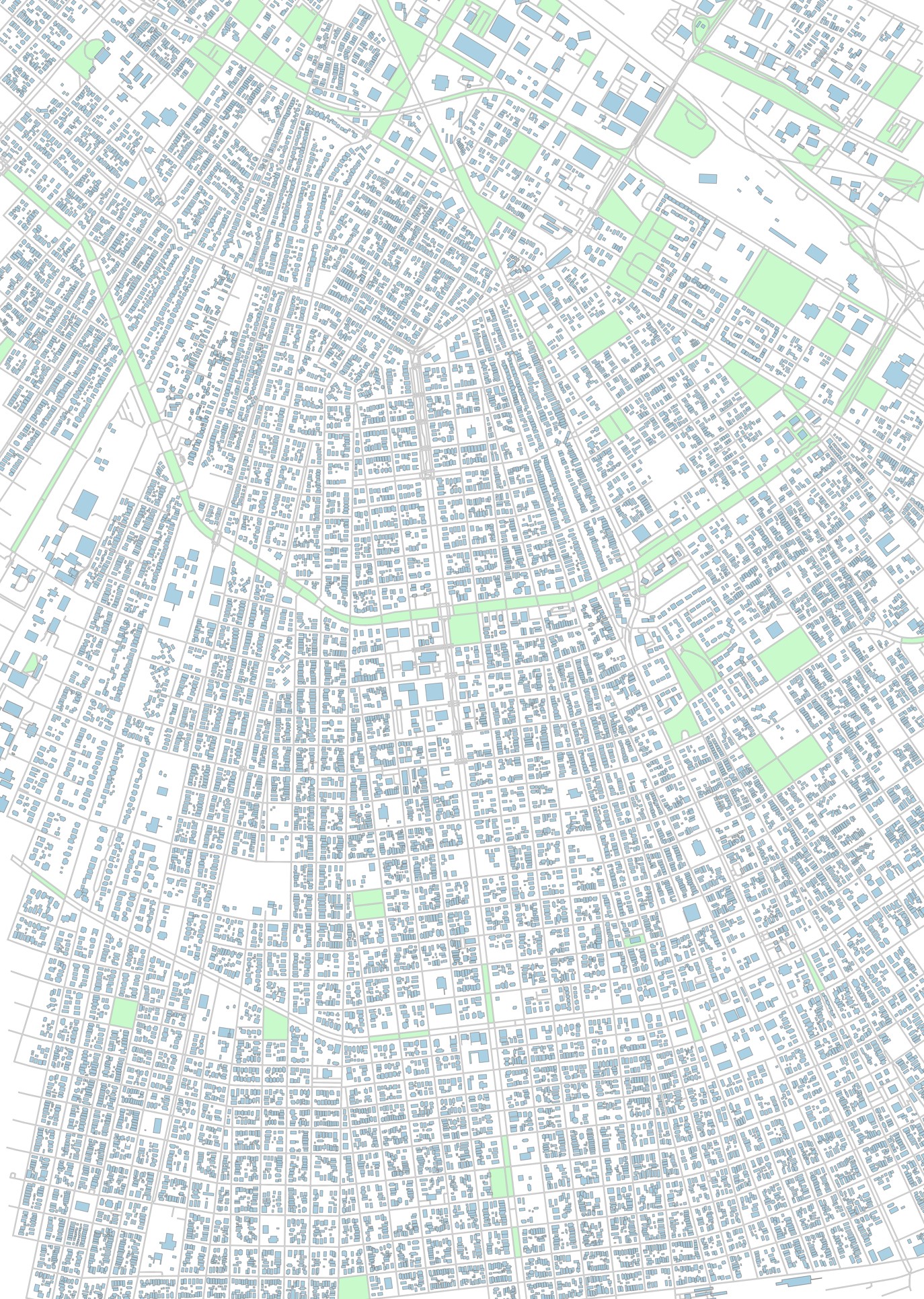}}
    
    \caption{\textbf{City-Scale Map.} We use the road networks from New Orleans to provide a city-scale generation using our method.}
  \label{fig:big_map}
\end{figure*}

\begin{figure*}
\centering

\frame{\includegraphics[width=0.95\textwidth]{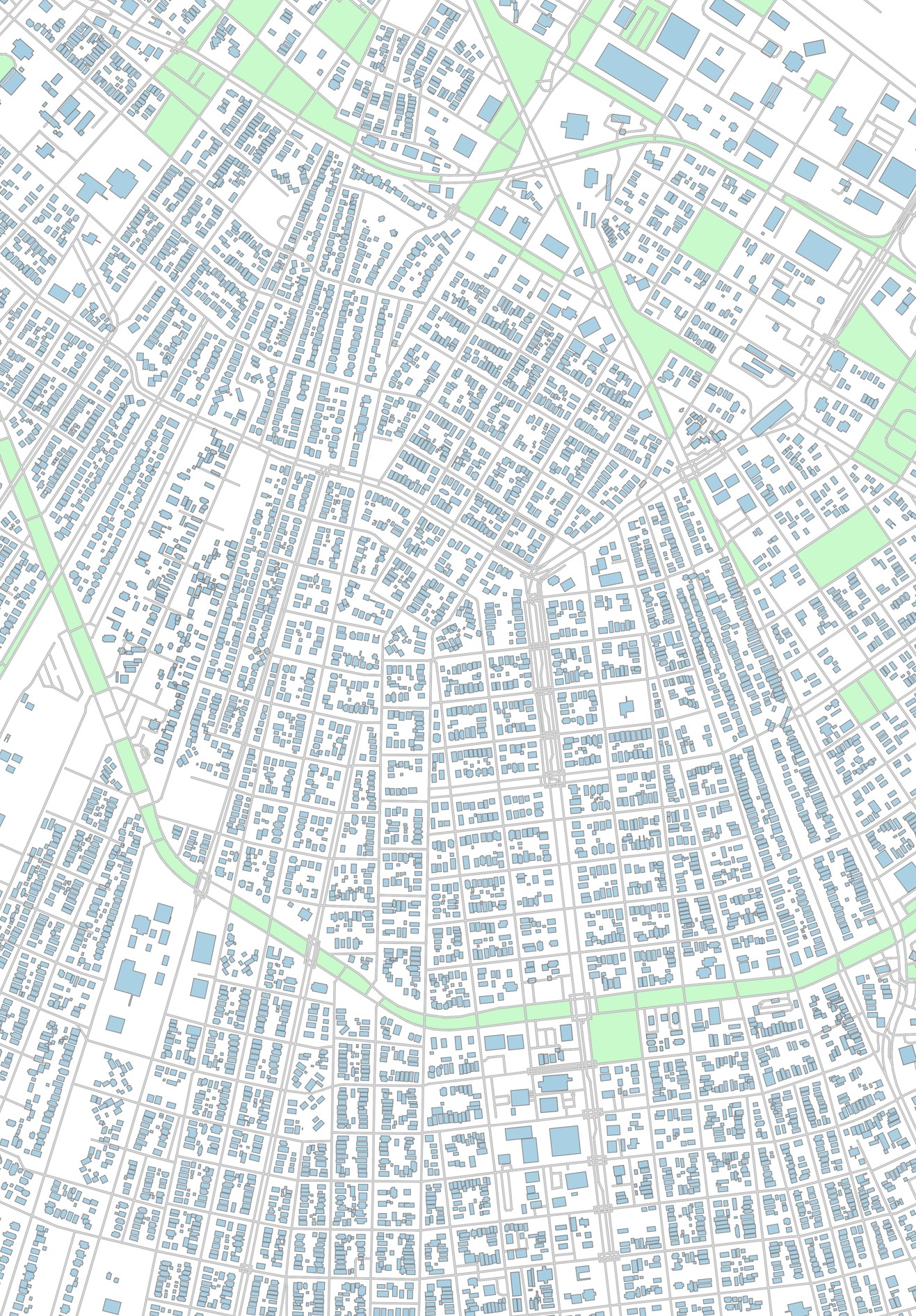}}
    
    \caption{\textbf{Zoom-in Map.} Zoom-in map of the upper right corner in Fig.~\ref{fig:big_map}.}
  \label{fig:crop_map}
\end{figure*}

In Fig.~\ref{fig:big_map} and its zoom-in in Fig.~\ref{fig:crop_map}, we present a city-scale generation given only road networks and 10 representative city blocks as small controlling priors (i.e., $<1\%$). Our method is able to generate realistic 2.5D urban layouts from road networks with plausible context harmonization from community-scale to city-scale.

\begin{figure*}[hbt!]
\centering
\setlength{\tabcolsep}{3pt}
\begin{tabular}{cccc}
   \includegraphics[width=0.23\textwidth]{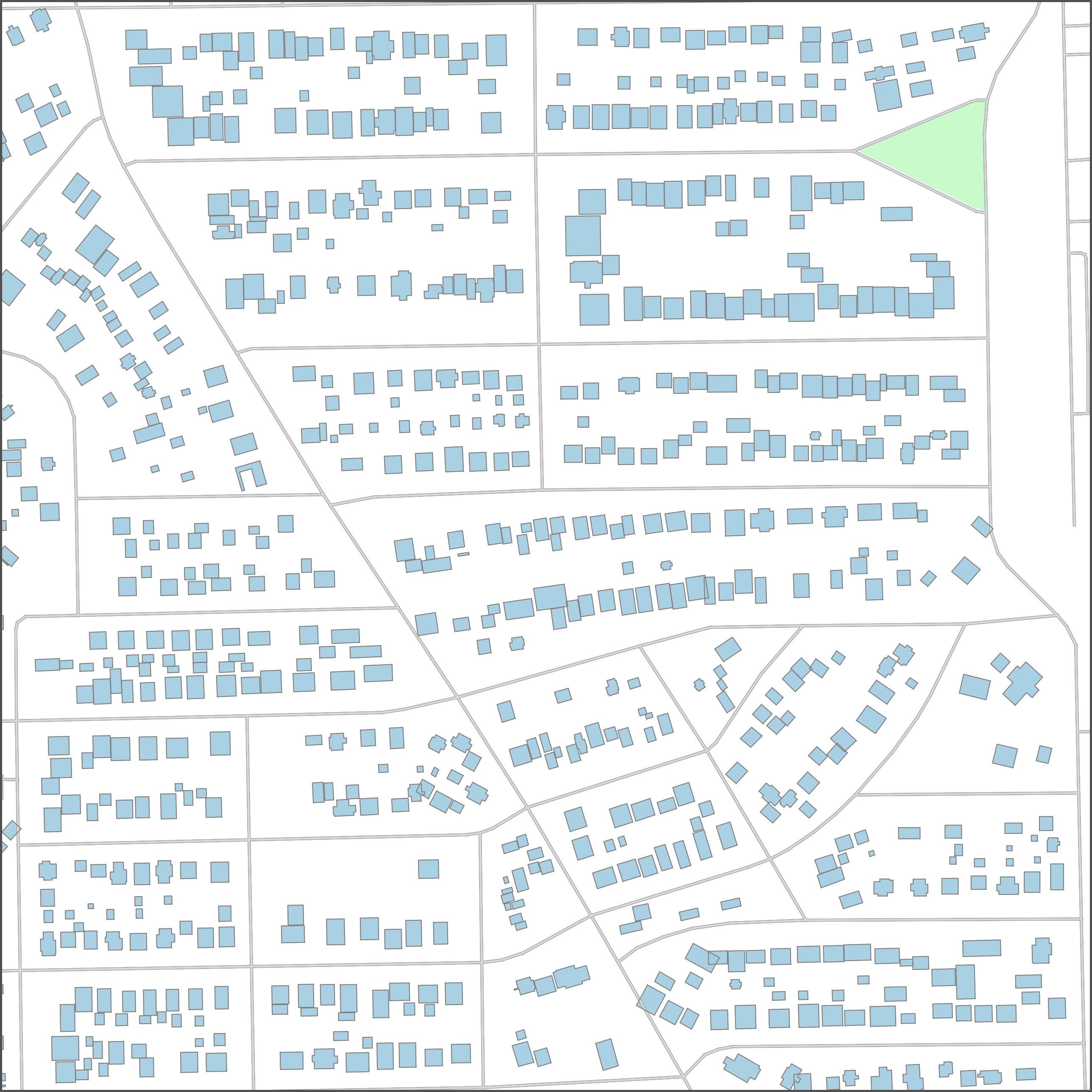} &
   \includegraphics[width=0.23\textwidth]{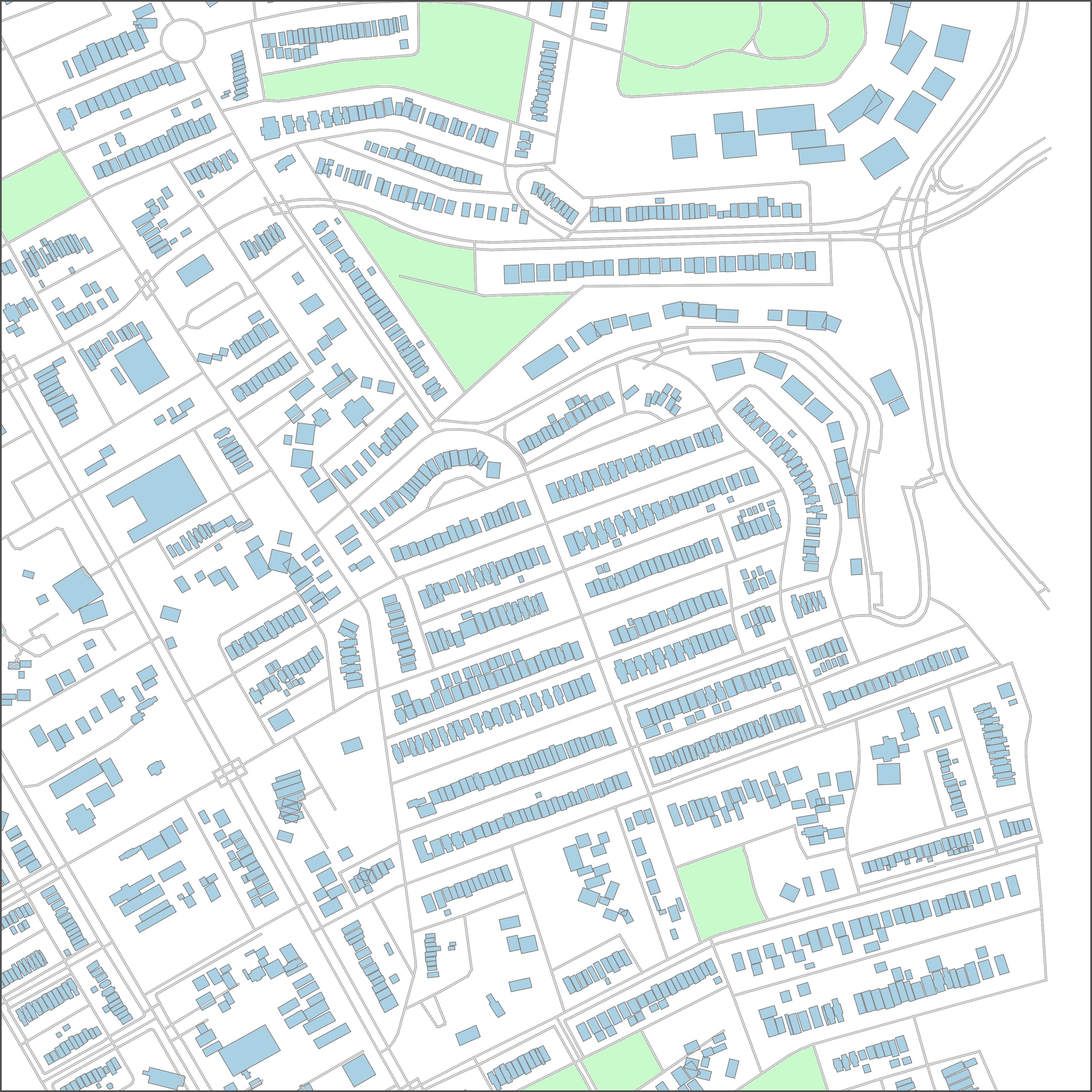} &
   \includegraphics[width=0.23\textwidth]{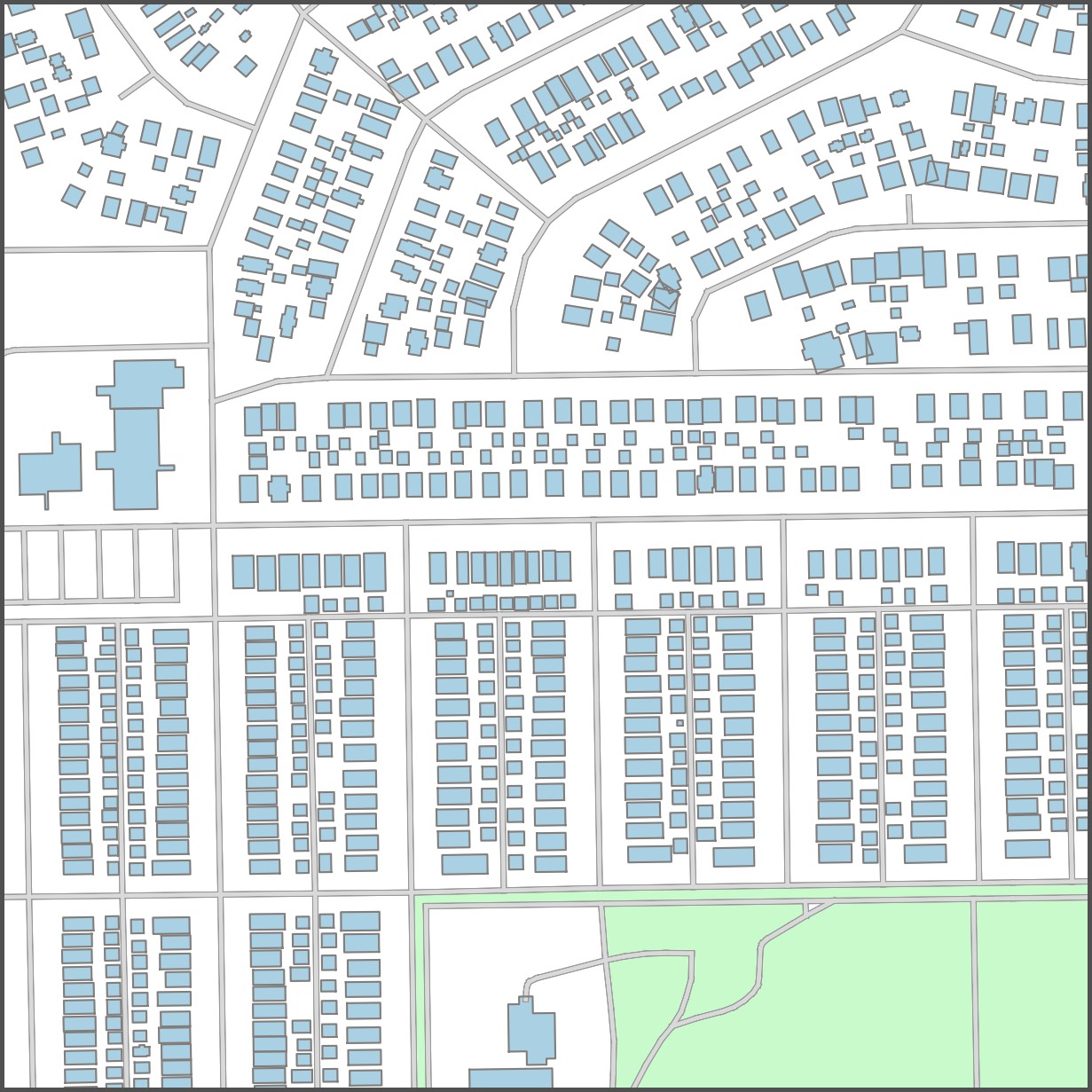} &    
   \includegraphics[width=0.23\textwidth]{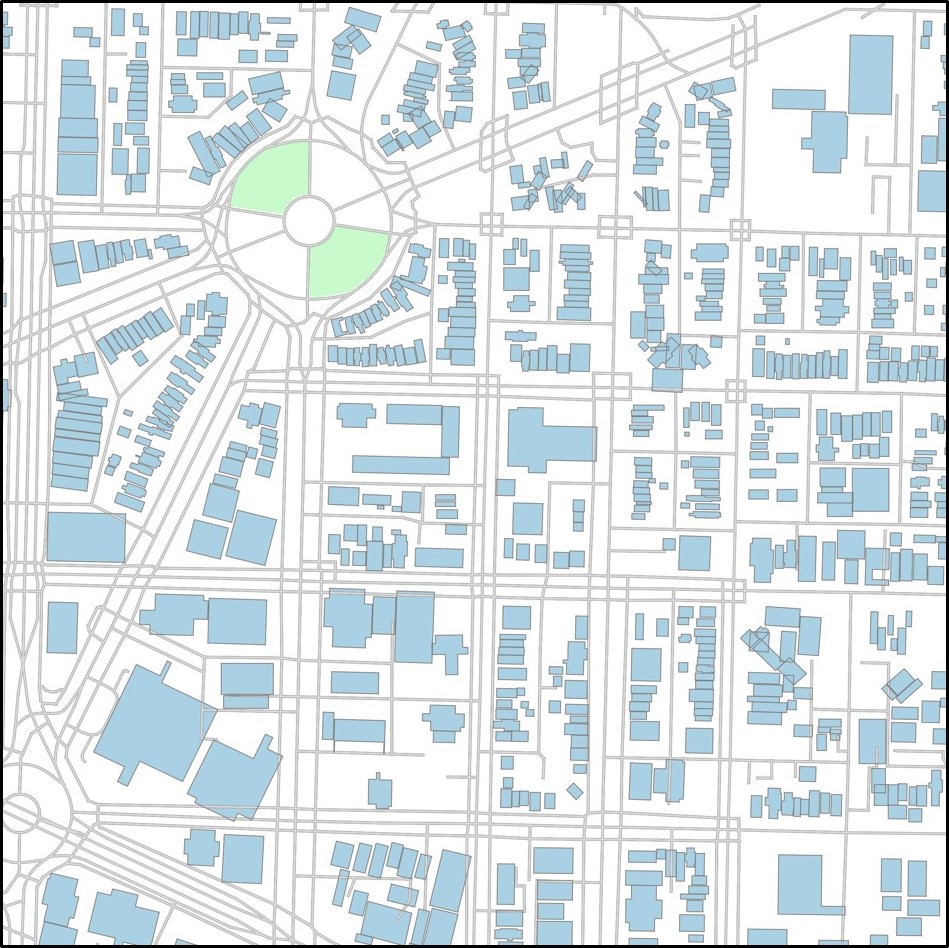} \\
     Atlanta & Baltimore & Chicago & D.C. \\ 
    \includegraphics[width=0.23\textwidth]{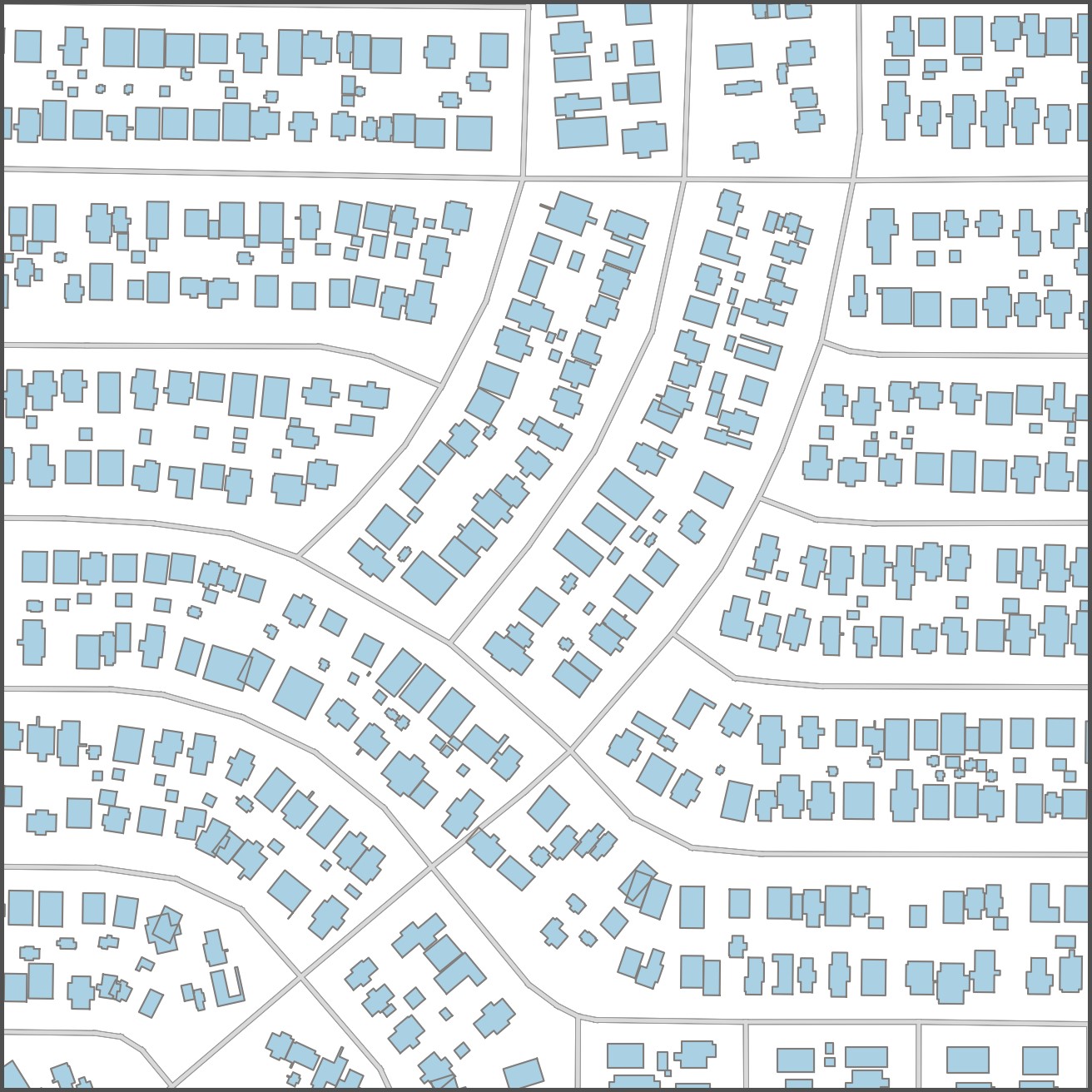} &
    \includegraphics[width=0.23\textwidth]{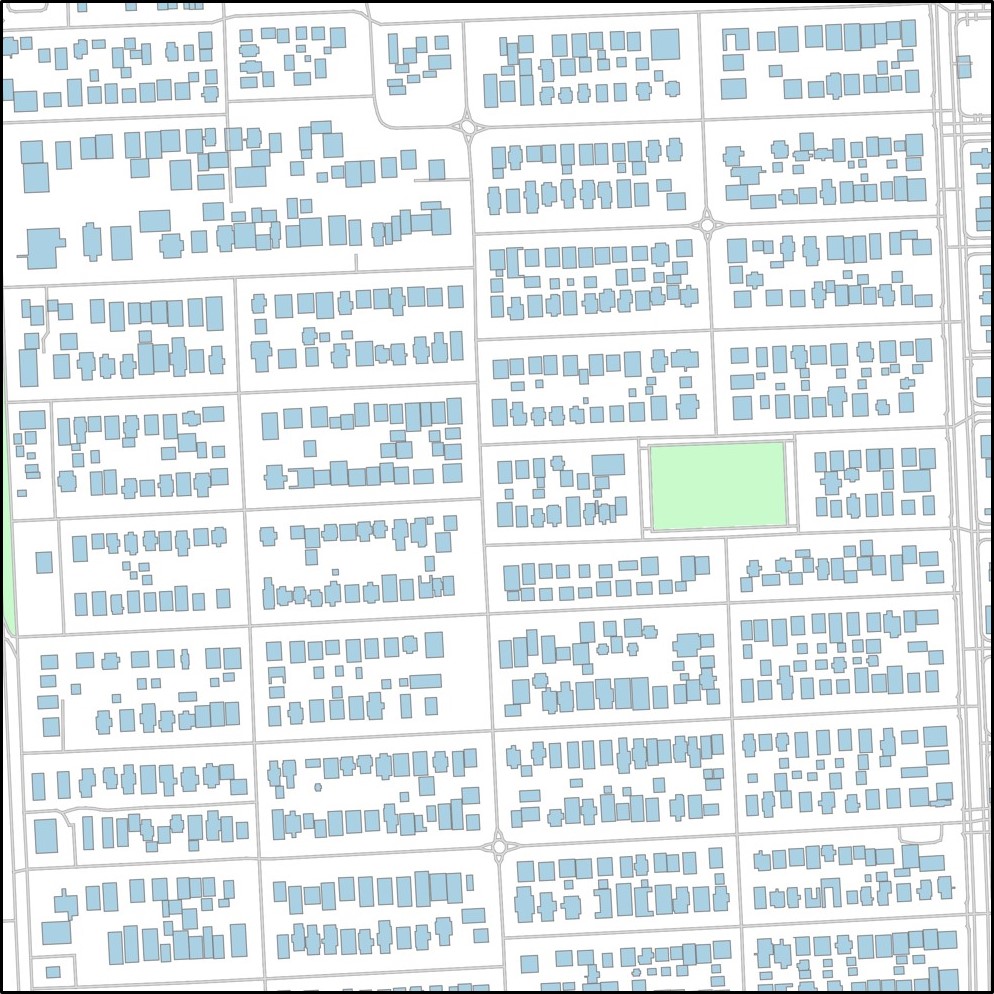} &
    \includegraphics[width=0.23\textwidth]{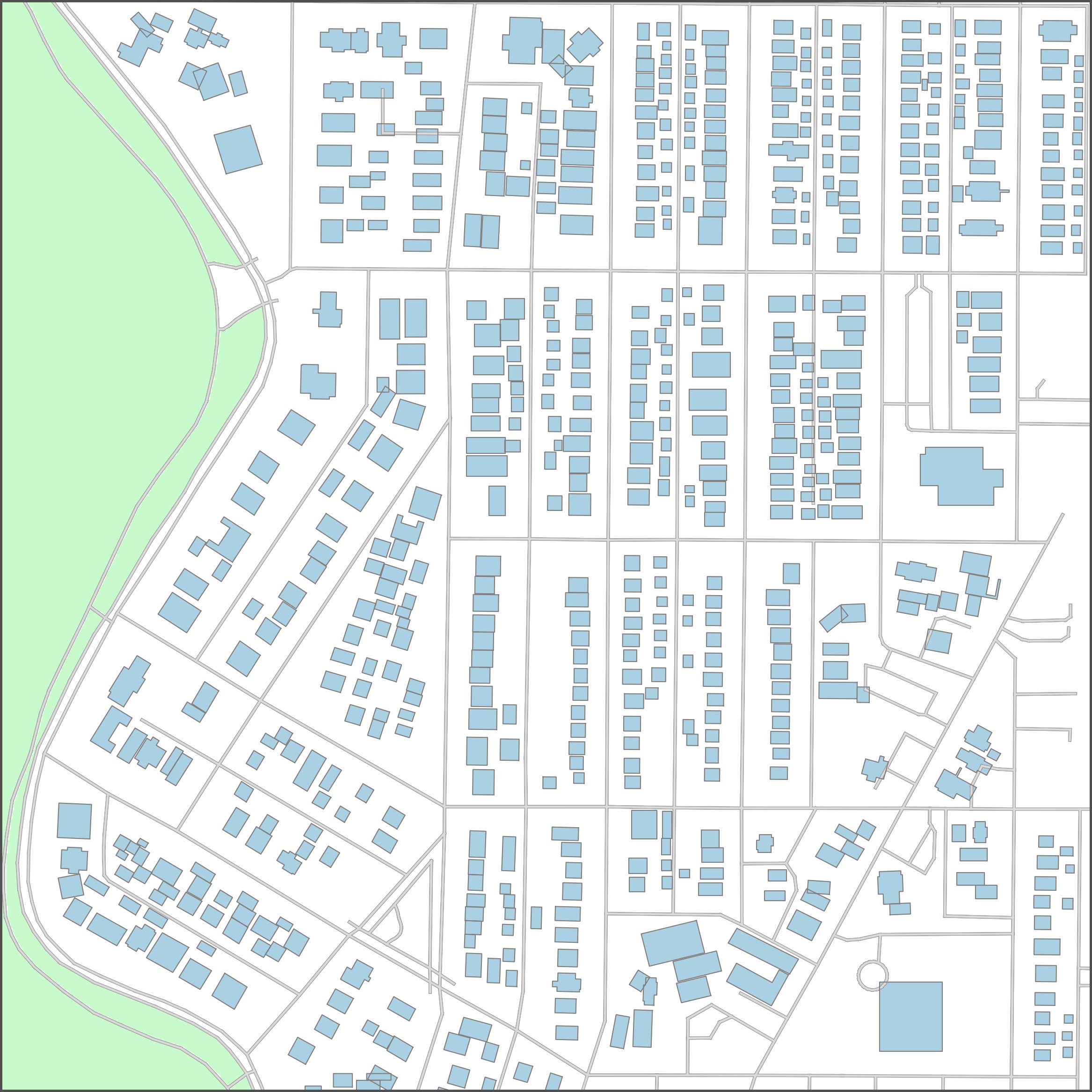} &
    \includegraphics[width=0.23\textwidth]{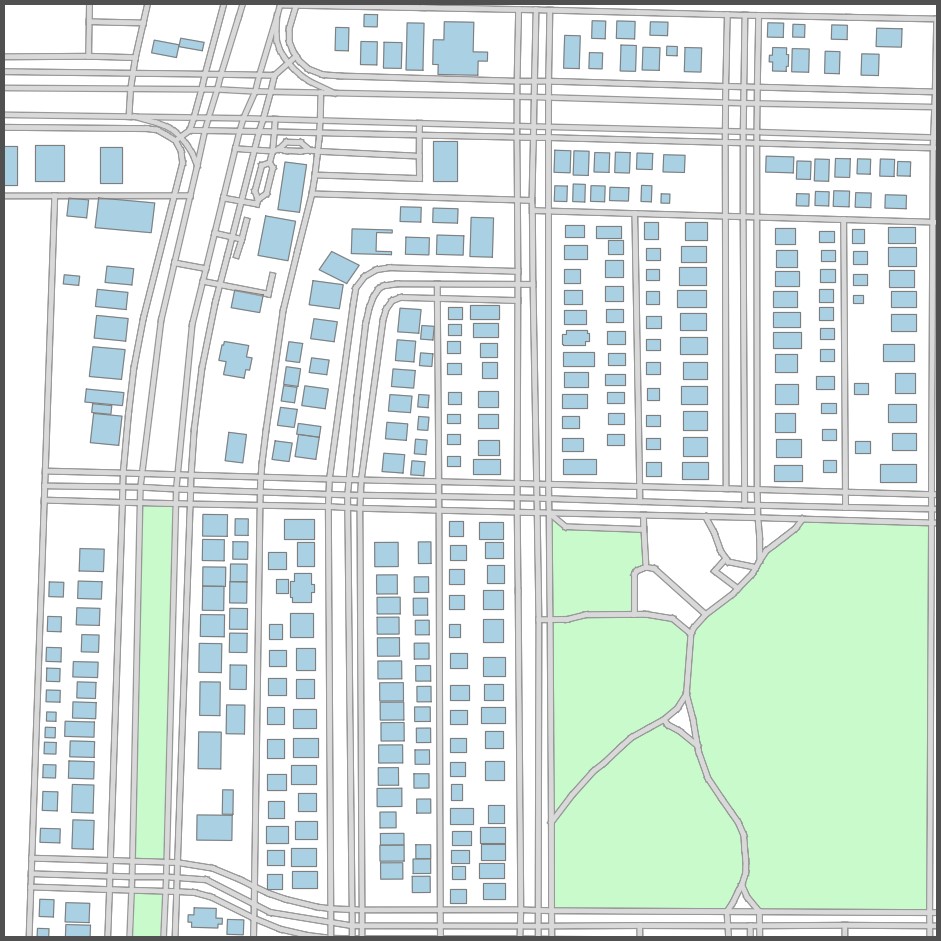} \\
     Los Angeles & Miami & Minneapolis & Milwaukee\\ 
    \includegraphics[width=0.23\textwidth]{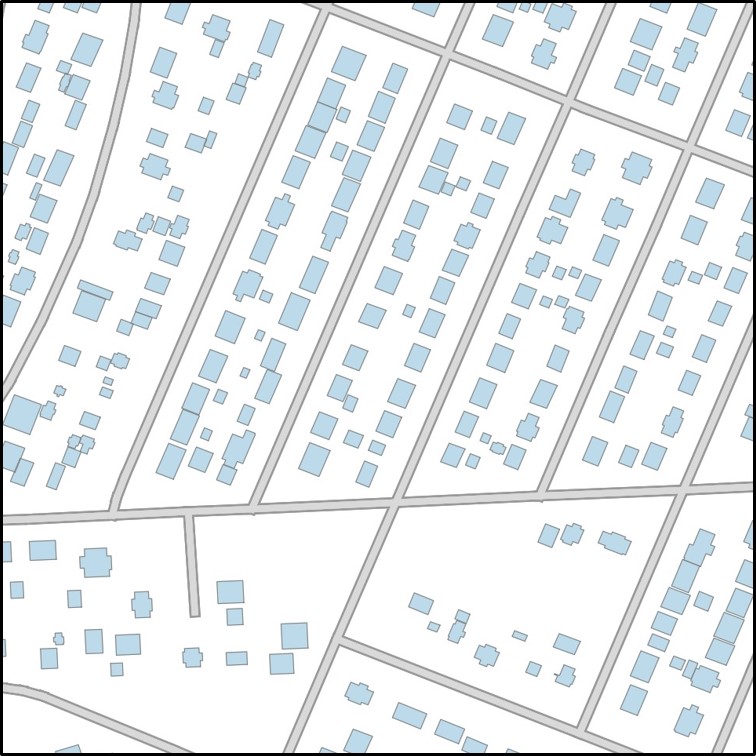} &
    \includegraphics[width=0.23\textwidth]{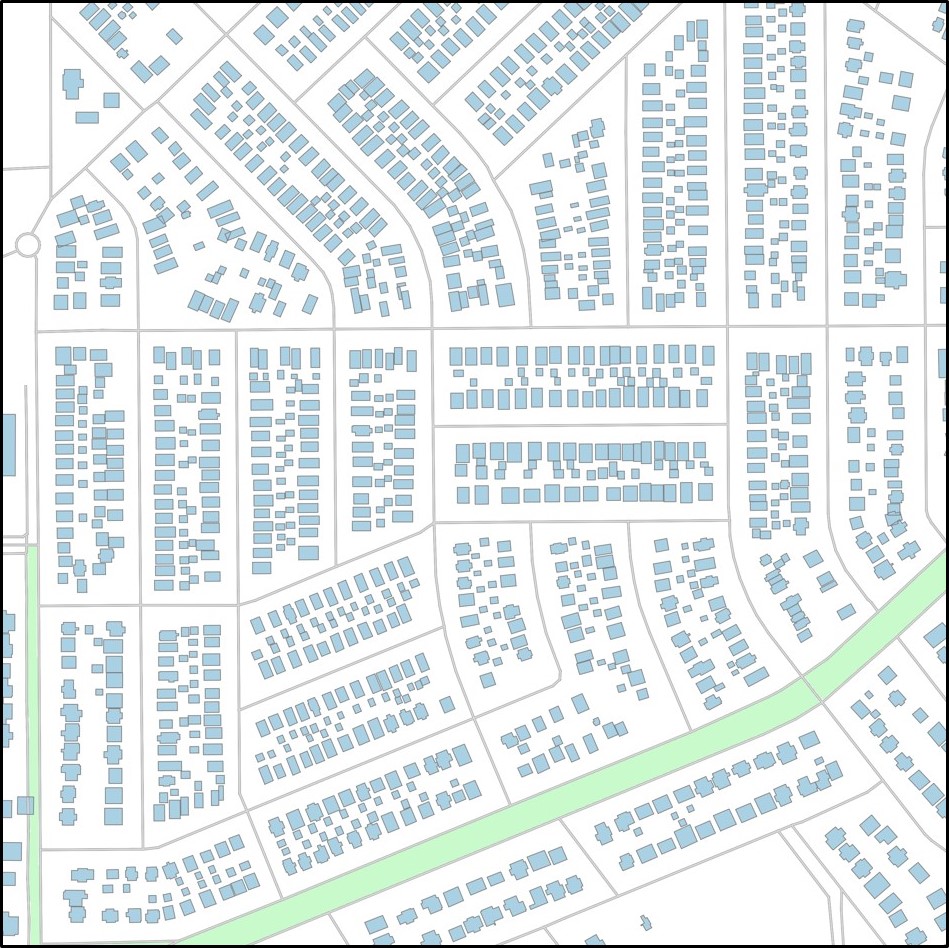} &
    \includegraphics[width=0.23\textwidth]{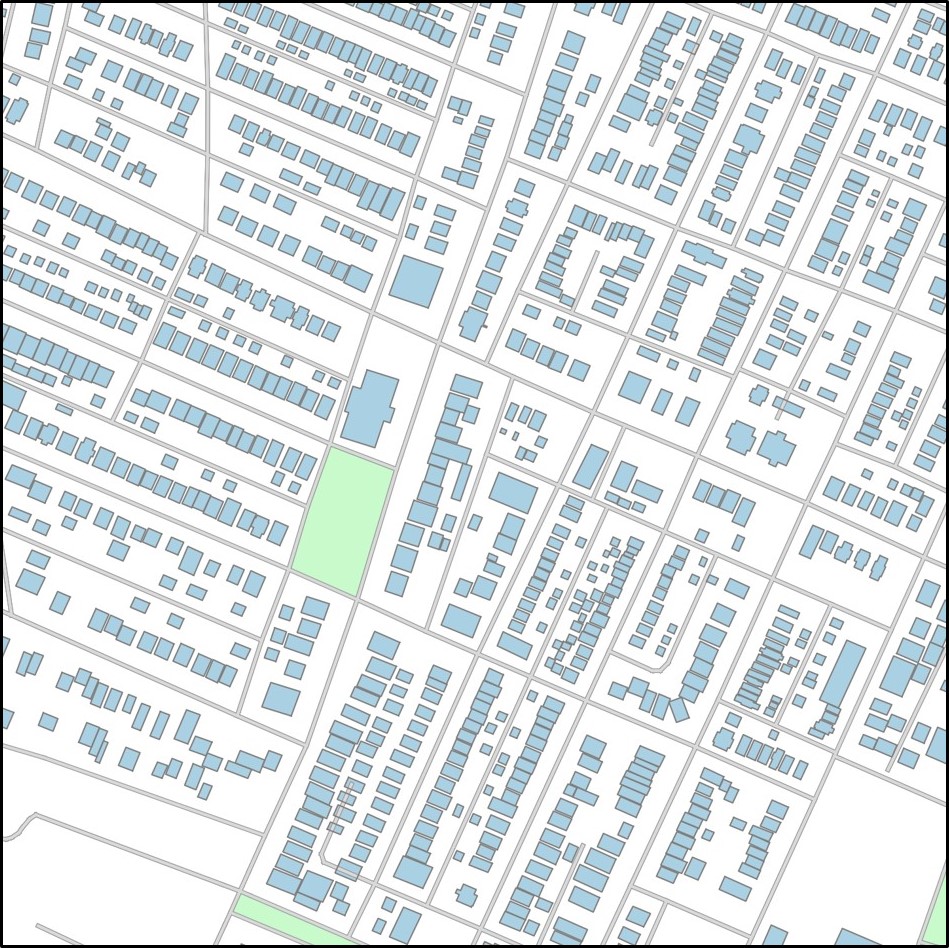} &
    \includegraphics[width=0.23\textwidth]{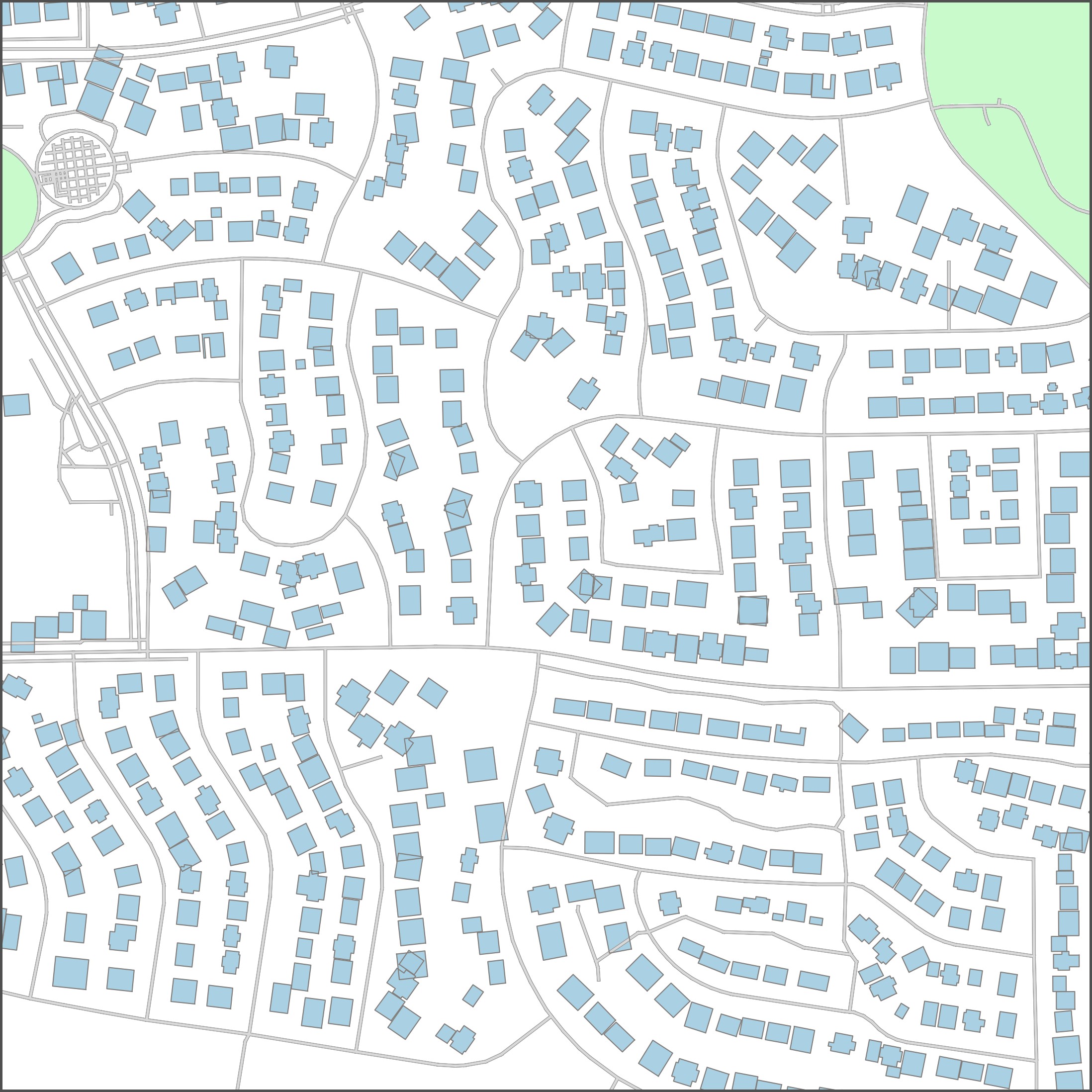} \\
     Norfolk & New York City & Pittsburgh & Portland\\ 
    \includegraphics[width=0.23\textwidth]{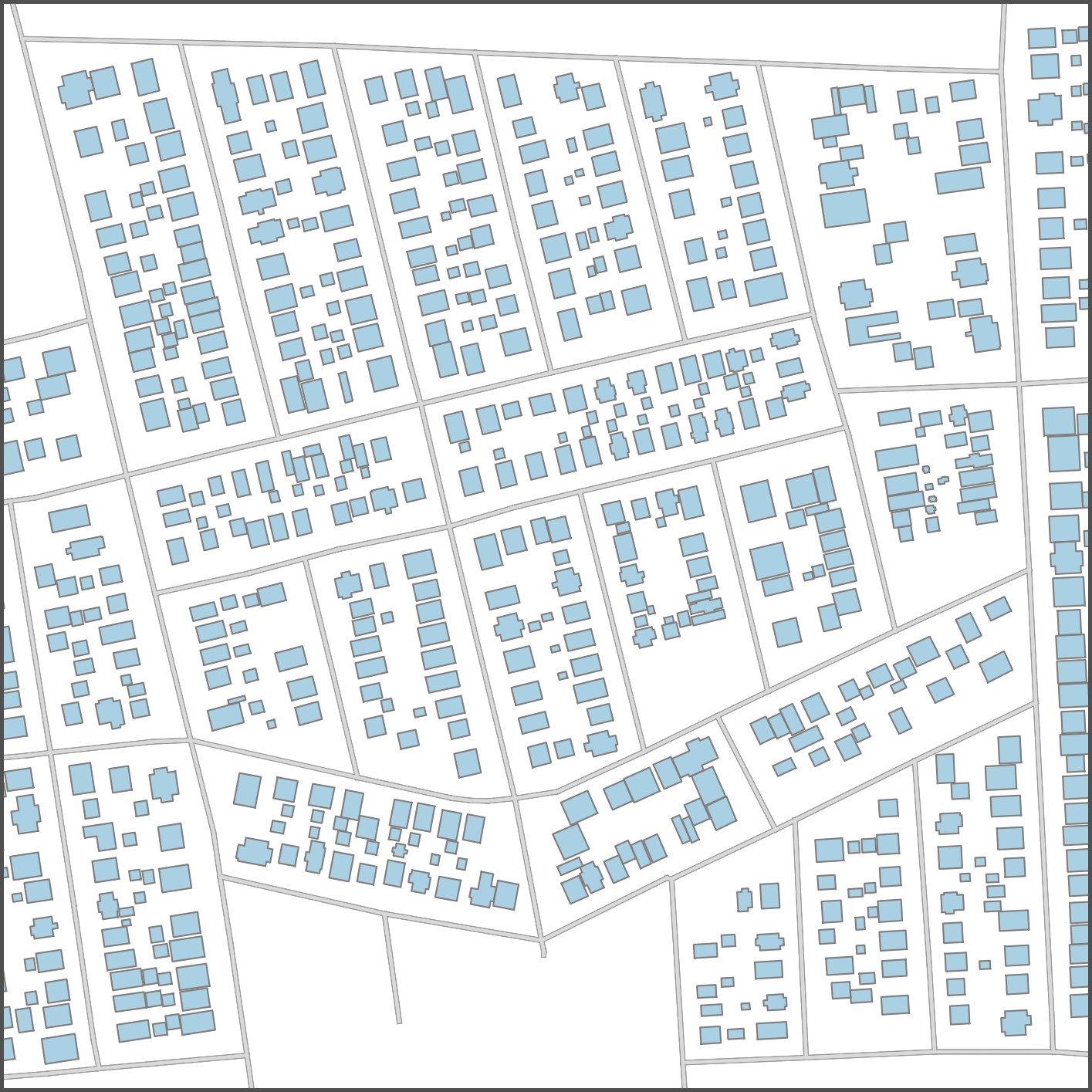} &
    \includegraphics[width=0.23\textwidth]{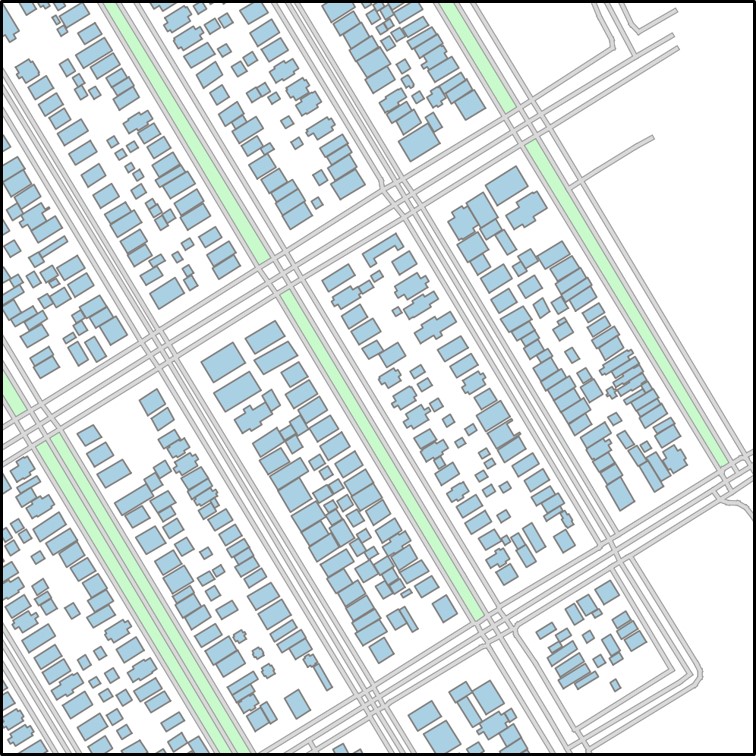} &
    \includegraphics[width=0.23\textwidth]{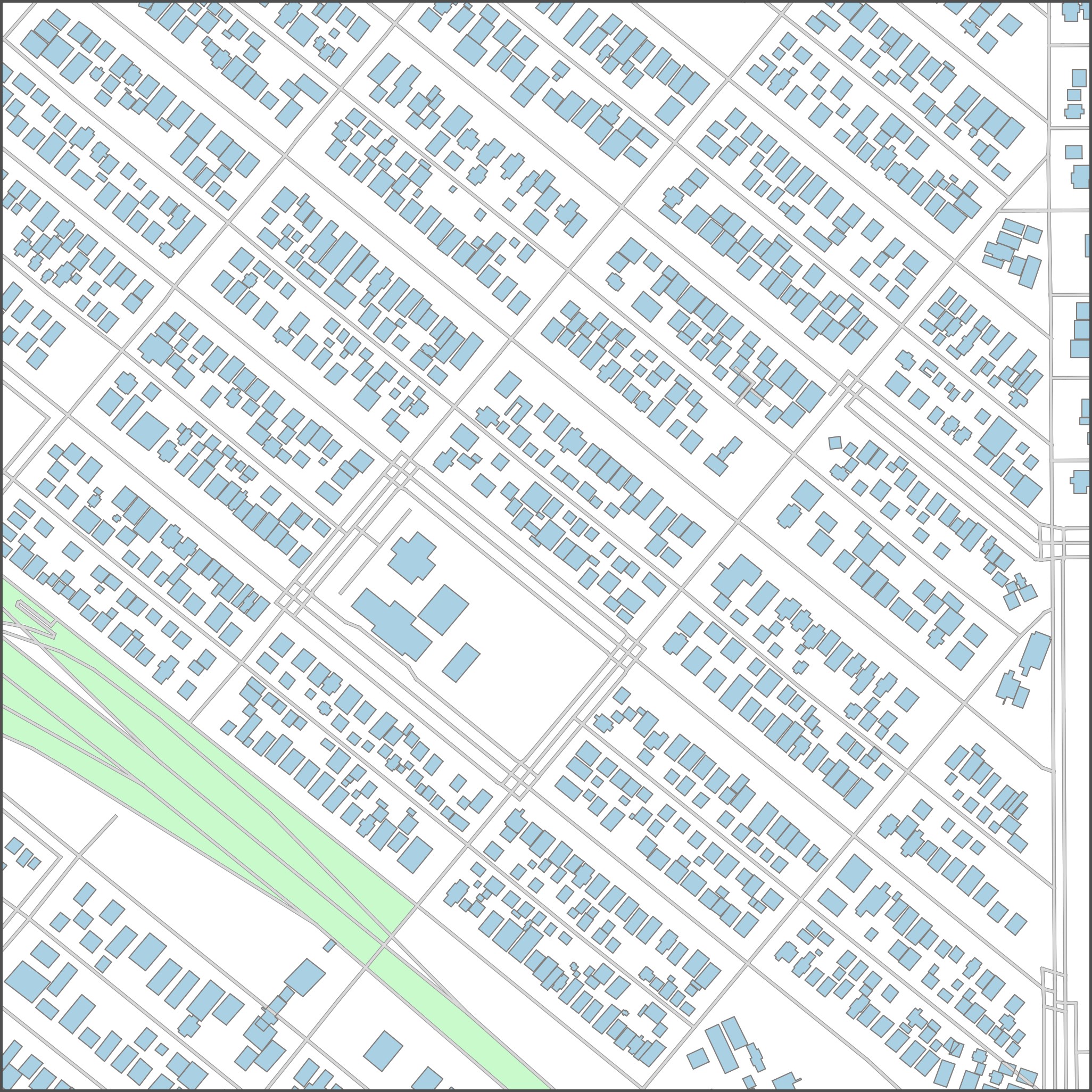} &
    \includegraphics[width=0.23\textwidth]{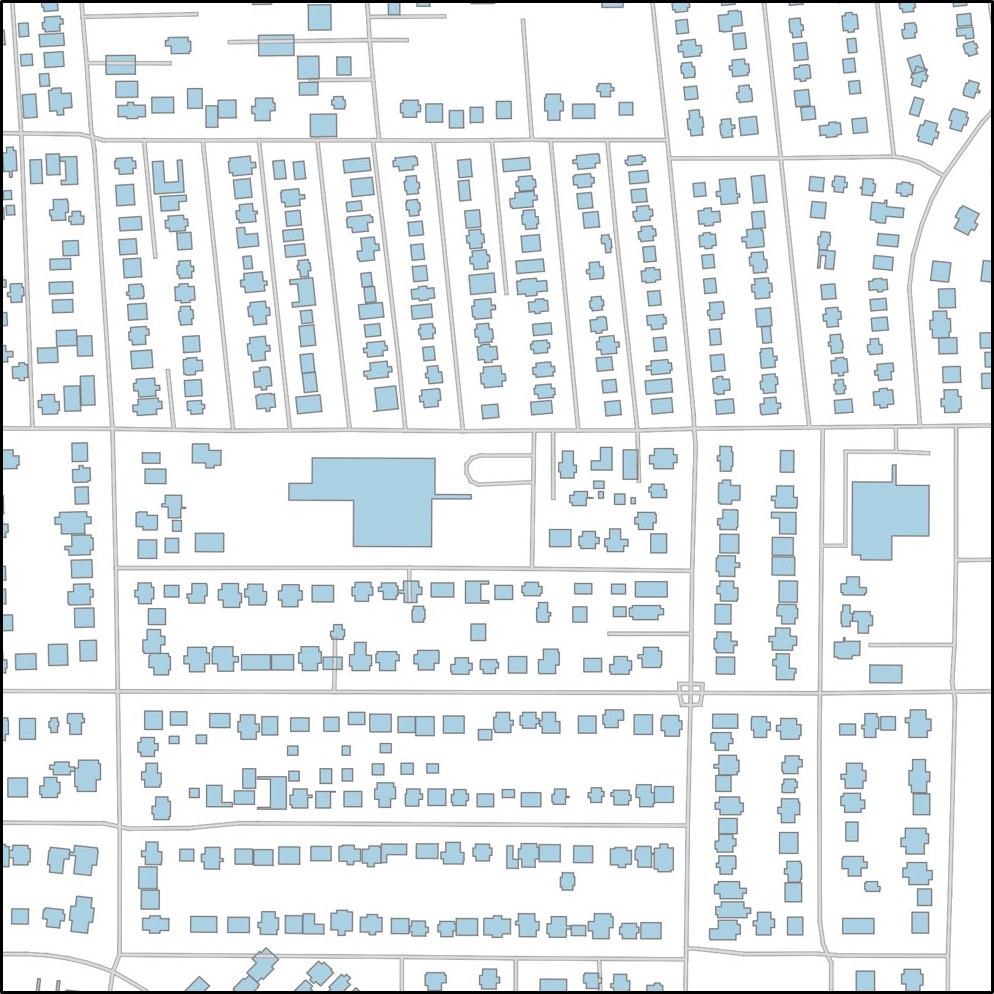} \\
     Providence & San Jose & San Diego & Seattle\\ 
\end{tabular}%
    \caption{\textbf{Example Generations across the U.S.} Given arbitrary road networks across 16 diverse cities, our method generates realistic urban layouts with plausible context harmonization.}
  \label{fig:rand gen}
\end{figure*}

Moreover, in Fig.~\ref{fig:rand gen}, we provide cropped generated examples for another 16 major cities across the U.S. It proves the robustness and generalization of our methods among heterogeneous city context styles.


\subsubsection{Comments on alternative approaches.} For all the comparisons in Sec. 4.2 of the main paper, we generate the result of each approach without post-processing or human-in-the-loop refinement. However, for some evaluation metrics, the metric value depends on the particular instance that was generated. Our method generates its output using our priority-based scheduling in $T$ iterations, with $T=12$ being a reasonable number. Hence, for fairness we also select the best results from $T$ independent generations for each alternative method.

Further, we tested other pixel-based approaches (e.g.\cite{lin2021infinitygan}) apart from SDXL~\cite{podell2023sdxl} (see Tab.1 and Fig.5 in the main paper). But none of them provide comparable results. Poor generation quality by InfinityGAN~\cite{lin2021infinitygan} is also observed by~\cite{xie2023citydreamer}.

\begin{figure*}[htb]
\centering

\includegraphics[width=\textwidth]{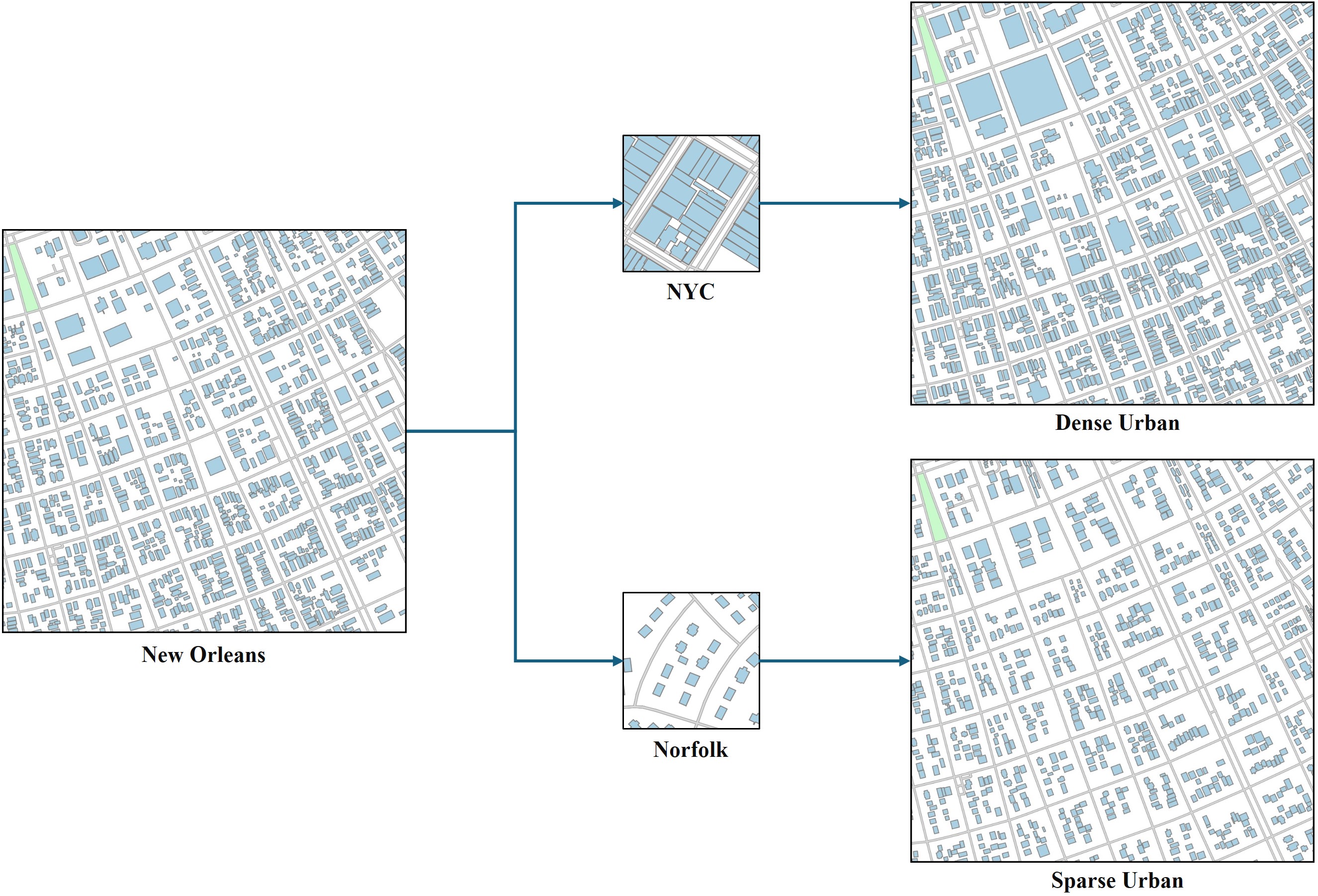}
    
    \caption{\textbf{City-scale Semantic Manipulation.} We use \textit{super nodes} to influence the generation of a New Orleans layout by using either dense building layout style from New York City or sparse urban style from Norfolk. See text for details.}
  \label{fig:semantic}
\end{figure*}

\begin{table}[h]
\small
\centering
  \caption{\textbf{GMAE-based Prediction.} Our GMAE combined with a conventional classifier (e.g. SVM or XGBoost) can be used to predict whether a set of city blocks corresponds to an advantaged and disadvantaged economic/social/environmental group.}

\resizebox{\columnwidth}{!}{\begin{tabular}{@{}llc@{}}
    \toprule
   Abbrv. & Metric Full Name & Best Acc. $\%$  \\
    \midrule
    DSF\_PFS & Diesel particulate matter exposure (percentile) & \textbf{89.76} \\
    EBF\_PFS & Energy burden (percentile) & \textbf{84.26}\\
    LMI\_PFS & Low median household income as a percent of area median income (percentile) & \textbf{83.62} \\
    LLEF\_PFS & Low life expectancy (percentile)  &  \textbf{80.30}\\
    EBLR\_PFS & Expected building loss rate (Natural Hazards Risk Index) (percentile) & \textbf{80.13} \\
    LPF\_PFS & Percent pre-1960s housing (lead paint indicator) (percentile) &  \textbf{79.83} \\
    HBF\_PFS & Housing burden (percent) (percentile) & \textbf{76.82} \\
    EPLR\_PFS & Expected population loss rate (Natural Hazards Risk Index) (percentile) & \textbf{75.82} \\
    P100\_PFS & Percent of individuals $< 100\%$ Federal Poverty Line (percentile) & \textbf{75.46}  \\
    FLD\_PFS & Share of properties at risk of flood in 30 years (percentile) & \textbf{75.46} \\
    TF\_PFS & Traffic proximity and volume (percentile) & \textbf{75.15} \\
    
    \bottomrule
\end{tabular}}

\label{tab_indicator}
\end{table}%

\section{Semantic Manipulation}
\label{supp_sec:semantic_manipulation}
Using the trained GMAE, semantic manipulation can be implemented by adding \textit{super nodes} with desired building layout styles to the original city graph. The added super nodes have a dense connectivity to neighboring communities and will influence the generation pattern of GMAE to produce a fused style. The semantic manipulation extent can be tuned by modifying connection degrees or edge distances attributes of the added super nodes. The users may customize their manipulations for broad "what-if" scenarios for urban planning, meteorology simulation, and game design applications.

In Fig.~\ref{fig:semantic}, we manipulate generated New Orleans with either dense urban priors from New York City or sparse urban priors from Norfolk. Our GMAE generates "fused" urban layouts guided by given priors. Meanwhile, the realism and diversity are kept.

\section{GMAE-based Socio-economic Metric Prediction}
\label{supp_sec:socio_eco}
We find that the 2.5D geometry of building layouts in a given city block correlates well with social, economic, and climate metrics of the same block. In particular, the encoder of our trained GMAE may act as a feature extractor for a binary classification of city blocks as advantaged or disadvantaged groups. Hence, we can also use the GMAE for socio-economic metric estimation without requiring tedious customized surveys, and potentially leading to broader social applications for policy making. This prediction is analogous to the "linear probe" in the representation learning community (e.g. CLIP~\cite{radford2021learning}). The image/text features extracted by a pretrained encoder (weights frozen) are directly fed into a trainable linear MLP adapting downstream tasks (e.g. classification).

We make use of the social-economic metric dataset from the Climate and Economic Justice Screening Tool (CEJST)~\cite{cjest}. This dataset provides census-tract-level data for more than 100 metrics. We calculate the values of those metrics for each city block, and label the city blocks in the lower 1st percentile as the disadvantaged group, and those in the higher 99th percentile as the advantaged group.

In this study, we utilize the extracted node features $F$ (see Fig.3 in main paper) to train a machine learning classifier. The classifier is trained to distinguish the aforementioned advantaged and disadvantaged groups. The performance of this binary classification, using SVM and XGBoost, is showed in Tab.~\ref{tab_indicator}. We report the diverse set of 11 metrics with higher than $75\%$ accuracy. 

\newpage

\section{Running time and model parameters}
\label{supp_sec:add_ablation}

Under the same settings as main paper, below we report model parameter size and average inference time per city block by a single RTX A5000. The only exception is for SDXL~\cite{podell2023sdxl} where the time is for generating one $1024\times1024$ resolution image by 20 sampling steps (typically covering dozens of blocks). Our method uses 12-iteration sampling. VTN~\cite{arroyo2021variational} and SDXL (even if normalized to single block inference time) are quite slow. LayoutDM~\cite{inoue2023layoutdm} (sampled by default 100 steps) is slower than our method, and GlobalMapper is slightly faster. However, our method outperforms others (main paper Tab.1 and Fig.5). Our model parameter size is much smaller than SDXL, and is only marginally larger than the largest of others.

\begin{table}[h]
\footnotesize
\centering
\begin{tabular}{lcc}
    \toprule
   Method & Inference Time$\downarrow$ (ms) & Parameters\\
    \midrule
   $\star$SDXL~\cite{podell2023sdxl} & 5938.85 & 2.6B\\
   VTN~\cite{arroyo2021variational} & 3428.32 & 45.49M\\
   LayoutDM~\cite{inoue2023layoutdm}  & 20.24 & 12.41M\\
   GlobalMapper~\cite{he2023globalmapper} & 5.62 & 76.48M\\
    \midrule
   \textbf{Ours} & 7.94 & 89.87M \\
    \bottomrule
\end{tabular}
\label{tab:comparison_TIME}
\end{table}%

\end{document}